\documentclass{article}

\PassOptionsToPackage{usenames,dvipsnames}{xcolor}

\usepackage[accepted]{icml2023}

\usepackage{amsmath, amssymb, amsthm}
\usepackage{mathtools}
\usepackage{bbm}
\usepackage{dsfont}

\usepackage{enumitem}

\usepackage{graphicx}
\usepackage{booktabs, array}
\usepackage{multirow}
\usepackage{subcaption}

\newsavebox\CBox

\newcommand{\tbf}[1]{\sbox\CBox{#1}\resizebox{\wd\CBox}{\ht\CBox}{\textbf{#1}}}
\newcommand{\tpm}[1]{\small{$\pm{\, #1}$}}
\newcommand{\umet}[1]{#1 ($\uparrow$)}
\newcommand{\dmet}[1]{#1 ($\downarrow$)}
\newcommand{\relv}[1]{$\times$ #1}

\usepackage{ifthen}
\newcommand{\tv}[3][n]{%
    \ifthenelse{\equal{#1}{n}}{%
        \ifthenelse{\equal{#3}{}}{#2}{#2 \tpm{#3}}%
    }{\ifthenelse{\equal{#1}{b}}{%
        \ifthenelse{\equal{#3}{}}{\tbf{#2}}{\tbf{#2} \tpm{#3}}%
    }{%
        \textcolor{red}{ERROR}%
    }%
}}

\usepackage{listings}
\usepackage{fancyvrb}
\fvset{fontsize=\small}

\usepackage[colorlinks,linktoc=all,pagebackref=true]{hyperref}
\usepackage[all]{hypcap}
\hypersetup{citecolor=NavyBlue}
\hypersetup{linkcolor=NavyBlue}
\hypersetup{urlcolor=NavyBlue}
\usepackage[nameinlink,capitalise, noabbrev]{cleveref}
\creflabelformat{equation}{#2\textup{#1}#3}  % <- remove parenthesis from equations
\crefname{section}{\S}{\S\S}
\usepackage[acronym,nowarn,section,nogroupskip,nonumberlist]{glossaries}
\glsdisablehyper{}

\newacronym{de}{\textsc{de}}{Deep Ensemble}
\newacronym{bma}{\textsc{bma}}{Bayesian Model Averaging}
\newacronym{mlp}{\textsc{mlp}}{Multilayer perceptron}
\newacronym{cnn}{\textsc{cnn}}{Convolutional Neural Network}
\newacronym{swa}{\textsc{swa}}{Stochastic Weight Averaging}
\newacronym{sgd}{\textsc{sgd}}{Stochastic Gradient Descent}
\newacronym{kd}{\textsc{kd}}{Knowledge Distillation}

\newcommand{\brgs}{$_\text{sm}$}
\newcommand{\brgm}{$_\text{md}$}

\newcommand{\calB}{{\mathcal{B}}}

\newcommand{\calD}{{\mathcal{D}}}

\newcommand{\calL}{{\mathcal{L}}}

\newcommand{\calO}{{\mathcal{O}}}

\newcommand{\calU}{{\mathcal{U}}}

\newcommand{\bbE}{\mathbb{E}}

\newcommand{\bbR}{\mathbb{R}}

\newcommand{\bsp}{\boldsymbol{p}}

\newcommand{\bsv}{\boldsymbol{v}}

\newcommand{\bsx}{\boldsymbol{x}}
\newcommand{\bsy}{\boldsymbol{y}}
\newcommand{\bsz}{\boldsymbol{z}}

\newcommand{\btheta}{{\boldsymbol{\theta}}}

\newcommand{\bphi}{{\boldsymbol{\phi}}}

\newcommand{\bpsi}{{\boldsymbol{\psi}}}
\newcommand{\bomega}{{\boldsymbol{\omega}}}

\newcommand{\bTheta}{\boldsymbol{\Theta}}

\newcommand{\bPhi}{\boldsymbol{\Phi}}
\newcommand{\bPsi}{\boldsymbol{\Psi}}

\theoremstyle{plain}%

\theoremstyle{definition}

\theoremstyle{remark}

\newcommand{\dee}{\mathrm{d}}

\DeclareMathOperator*{\argmax}{arg\,max}

\def\[#1\]{\begin{align}#1\end{align}}

\newcommand{\commentout}[1]{}

\newcommand{\bthetabe}{\btheta^{(\text{be})}}
\newcommand{\fext}{f^{(\text{ft})}}
\newcommand{\fcls}{f^{(\text{cls})}}
\newcommand{\typeone}{h^{(r)}}
\newcommand{\typetwo}{H^{(r)}}

\usepackage[textsize=tiny]{todonotes}

\icmltitlerunning{Traversing Between Modes in Function Space for Fast Ensembling}

\begin{document}

\twocolumn[
\icmltitle{Traversing Between Modes in Function Space for Fast Ensembling}

% It is OKAY to include author information, even for blind
% submissions: the style file will automatically remove it for you
% unless you've provided the [accepted] option to the icml2023
% package.

% List of affiliations: The first argument should be a (short)
% identifier you will use later to specify author affiliations
% Academic affiliations should list Department, University, City, Region, Country
% Industry affiliations should list Company, City, Region, Country

% You can specify symbols, otherwise they are numbered in order.
% Ideally, you should not use this facility. Affiliations will be numbered
% in order of appearance and this is the preferred way.
\icmlsetsymbol{equal}{*}

\begin{icmlauthorlist}
\icmlauthor{EungGu Yun}{equal,kaist,saige}
\icmlauthor{Hyungi Lee}{equal,kaist}
\icmlauthor{Giung Nam}{equal,kaist}
\icmlauthor{Juho Lee}{kaist,aitrics}
\end{icmlauthorlist}

\icmlaffiliation{kaist}{Kim Jaechul Graduate School~of~AI, Korea Advanced Institute of Science and Technology (KAIST), Daejeon, Korea}
\icmlaffiliation{saige}{Saige Research, Seoul, Korea}
\icmlaffiliation{aitrics}{AITRICS, Seoul, Korea}

\icmlcorrespondingauthor{EungGu Yun}{\mbox{eunggu.yun@kaist.ac.kr}}
\icmlcorrespondingauthor{Juho Lee}{juholee@kaist.ac.kr}

% You may provide any keywords that you
% find helpful for describing your paper; these are used to populate
% the "keywords" metadata in the PDF but will not be shown in the document
\icmlkeywords{Machine Learning, ICML, Deep Ensemble, Loss Landscape, Mode Connectivity}

\vskip 0.3in
]

% this must go after the closing bracket ] following \twocolumn[ ...

% This command actually creates the footnote in the first column
% listing the affiliations and the copyright notice.
% The command takes one argument, which is text to display at the start of the footnote.
% The \icmlEqualContribution command is standard text for equal contribution.
% Remove it (just {}) if you do not need this facility.

% \printAffiliationsAndNotice{}  % leave blank if no need to mention equal contribution
\printAffiliationsAndNotice{\icmlEqualContribution} % otherwise use the standard text.

\begin{abstract}
Deep ensemble is a simple yet powerful way to improve the performance of deep neural networks.
Under this motivation, recent works on mode connectivity have shown that parameters of ensembles are connected by low-loss subspaces, and one can efficiently collect ensemble parameters in those subspaces. 
While this provides a way to efficiently train ensembles, for inference, multiple forward passes should still be executed using all the ensemble parameters, which often becomes a serious bottleneck for real-world deployment. 
In this work, we propose a novel framework to reduce such costs. 
Given a low-loss subspace connecting two modes of a neural network, we build an additional neural network that predicts the output of the original neural network evaluated at a certain point in the low-loss subspace.
The additional neural network, which we call a ``bridge'', is a lightweight network that takes minimal features from the original network and predicts outputs for the low-loss subspace without forward passes through the original network. 
We empirically demonstrate that we can indeed train such bridge networks and significantly reduce inference costs with the help of bridge networks.
\end{abstract}

\section{Introduction}
\label{main:sec:introduction}

\gls{de}~\citep{lakshminarayanan2017simple} is a simple algorithm to improve both the predictive accuracy and the uncertainty calibration of deep neural networks, where a neural network is trained multiple times using the same data but with different random seeds.
Due to this randomness, the parameters obtained from the multiple training runs reach different local optima, called modes, on the loss surface~\citep{fort2019de}. These parameters represent a set of various functions that serve as an effective approximation for \gls{bma}~\citep{wilson2020bayesian}.

An apparent drawback of \gls{de} is that it requires multiple training runs. This cost can be huge especially for large-scale settings for which parallel training is not feasible. \citet{garipov2018loss} and \citet{draxler2018essentially} showed that the modes in the loss surface of a deep neural network are connected by relatively simple low-dimensional subspaces where every parameter in those subspaces retains low training error, and the parameters along those subspaces are good candidates for ensembling.
Based on this observation, \citet{garipov2018loss} and \citet{huang2017snapshot} proposed algorithms to quickly construct deep ensembles without having to run multiple independent training runs.

While the fast ensembling methods based on mode connectivity reduce training costs, they do not address another important drawback of \gls{de}; \textit{the inference cost}.
One should still execute multiple forward passes using all the parameters collected for the ensemble, and this cost often becomes critical for a real-world scenario,  where the training is done in a resource-abundant setting with plenty of computation time, but for the deployment, the inference should be done in a resource-limited environment.
For such settings, reducing the inference cost is much more important than reducing the training cost.

In this paper, we propose a novel approach to scale up \gls{de} by reducing the inference cost.
We start from an assumption; if two modes in an ensemble are connected by a simple subspace, we can predict the outputs corresponding to the parameters in the subspace using \emph{only the outputs computed from the modes.}
In other words, we can predict the outputs evaluated in the subspace without having to forward the actual parameters in the subspace through the network.
If this is indeed possible, for instance, given two modes, we can approximate an ensemble of three models consisting of parameters collected from three different locations (one from the subspace connecting two modes, and two from each mode) with only two forward passes and a small auxiliary forward pass.

We show that we can actually implement this idea using an additional lightweight network whose inference cost is relatively low compared to that of the original neural network.
This additional network, what we call a ``bridge network'', takes some features from the original neural network (e.g., features from the penultimate layer) and directly predicts the outputs computed from the connecting subspace.
In other words, the bridge network lets us travel between modes in the \emph{function space}.

We present two types of bridge networks depending on the number of modes involved in prediction, network architectures for bridge networks, and training procedures.
Through empirical validation on various image classification benchmarks, we show that (1) bridge networks can predict outputs of connecting subspaces quite accurately with minimal computation cost, and (2) \glspl{de} augmented with bridge networks can significantly reduce inference costs without big sacrifice in performance.

\begin{figure*}
    \vspace{2mm}
    \centering
    \includegraphics[width=0.30\textwidth]{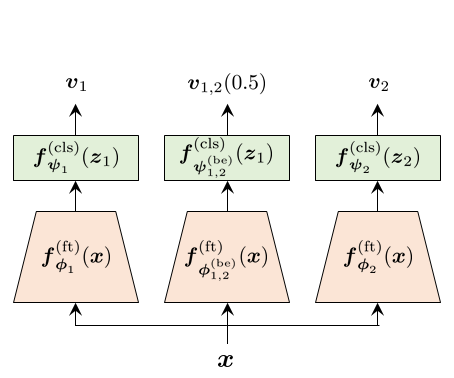}\label{fig:bezier_flow}
    \hspace{3mm}
    \includegraphics[width=0.30\textwidth]{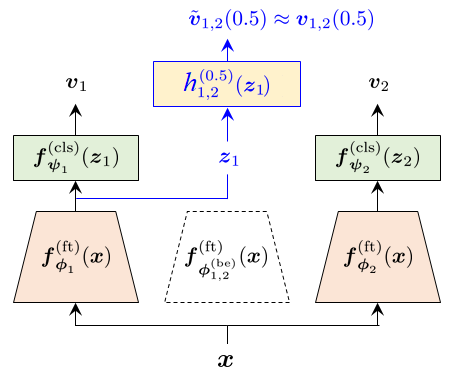}\label{fig:bridge_flow_type1}
    \hspace{3mm}
    \includegraphics[width=0.30\textwidth]{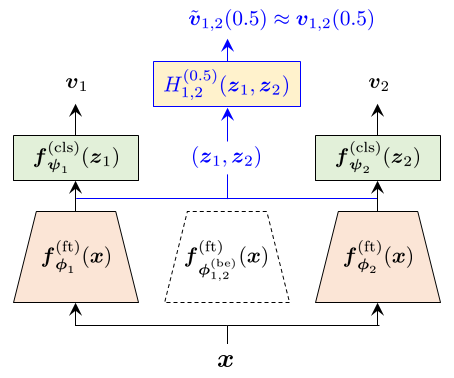}\label{fig:bridge_flow_type2}
    \caption{Ensembles with a Bezier curve (\textbf{left}), a type I bridge network (\textbf{center}), and a type II bridge network (\textbf{right}).}
    \label{fig:flow_comparison}
    \vspace{-3mm}
\end{figure*}

\section{Preliminaries}
\label{main:sec:preliminaries}

\subsection{Problem setup}
In this paper, we discuss the $K$-way classification problem taking $D$-dimensional inputs. 
A classifier is constructed with a deep neural network $f_{\btheta}: \bbR^{D}\to\bbR^{K}$ which is
decomposed into a \emph{feature extractor} $\fext_\bphi: \bbR^{D}\to\bbR^{D_\text{ft}}$ and a \emph{classifier} $\fcls_{\bpsi}: \bbR^{D_\text{ft}}\to \bbR^K$, i.e., $f_{\btheta}(\bsx) =  \fcls_\bpsi\circ \fext_\bphi(\bsx)$. Here, $\bphi\in \bPhi$ and $\bpsi \in \bPsi$ denote the parameters for the feature extractor and classifier, respectively, $\btheta = (\bphi, \bpsi) \in \bTheta$, and $D_\text{ft}$ is the dimension of the feature.
An output from the classifier corresponds to a class probability vector.

\subsection{Finding low-loss subspaces}
While there are few low-loss subspaces that are known to connect modes of deep neural networks, in this paper, we focus on \emph{Bezier curves} as suggested in \citep{garipov2018loss}. Let $\btheta_i$ and $\btheta_j$ be two parameters (usually corresponding to modes) of a neural network. The quadratic Bezier curve between them is defined as
\[
\Big\{ (1-r)^2\btheta_i + 2r(1-r)\bthetabe_{i,j} + r^2\btheta_j \,|\, r \in [0,1] \Big\},
\]
where $\bthetabe_{i,j}$ is a \emph{pin-point} parameter characterizing the curve. Based on this curve paramerization, a low-loss subspace connecting $(\btheta_i, \btheta_j)$ is found by minimizing the following loss w.r.t. $\bthetabe_{i,j}$,
\[
\int_0^1 \calL\Big(\bthetabe_{i,j}(r)\Big) \dee r,
\]
where $\bthetabe_{i,j}(r)$ denotes the parameter at the position $r$ of the curve,
\[
\bthetabe_{i,j}(r) = (1-r)^2\btheta_i + 2r(1-r)\bthetabe_{i,j} + r^2\btheta_j,
\]
and $\calL:\bTheta \to \bbR$ is the loss function evaluating parameters (e.g., cross entropy). Since the integration above is usually intractable, we instead minimize the stochastic approximation:
\[
\bbE_{r \sim \calU(0, 1)} \left[
    \calL\left(\bthetabe_{i,j}(r) \right)
\right],
\]
where $\calU(0,1)$ is the uniform distribution on $[0, 1]$. For a more detailed procedure of Bezier curve training, please refer to \citet{garipov2018loss}.

\subsection{Ensembles with Bezier curves}
Let $\{ \btheta_1, \dots, \btheta_{m}\}$ be a set of parameters independently trained as a deep ensemble. Then, for each pair $(\btheta_i, \btheta_j)$, we can construct a low-loss Bezier curve. Since all of the parameters along those Bezier curves achieve low loss, we can add them to the ensemble for improved performance. For instance, choosing $r=0.5$, we can collect $\bthetabe_{i,j}(0.5)$ for all $(i,j)$ pairs, and construct an ensembled predictor as
\[
\frac{1}{m + \binom{m}{2}}\bigg(\sum_{i=1}^m f_{\btheta_i}(\bsx) + \sum_{i<j} f_{\bthetabe_{i,j}(0.5)}(\bsx)\bigg).
\]
While this strategy provide an effective way to increase the number of ensemble members, for inference, an additional $\calO(m^2)$ number of forward passes are required. Our primary goal in this paper is to reduce this additional cost by bypassing the direct forward passes with $\bthetabe_{i,j}(r)$.

\section{Main contribution}
\label{main:sec:methods}

In this section, we present a novel method that directly predicts the outputs of neural networks evaluated at parameters on Bezier curves without actual forward passes with them.

\subsection{Bridge networks}
Let us first recall our key assumption stated in the introduction; if two modes in an ensemble are connected by a simple low-loss subspace (Bezier curve), then we can predict the outputs corresponding to the parameters on the subspace using only the information obtained from the modes. The intuition behind this assumption is that, since the parameters are connected with a simple curve, the corresponding outputs may also be connected via a relatively simple mapping, which is far less complex than the original neural network. If such mapping exists, we may learn them via a lightweight neural network.

More specifically, let $\bsz_i := \fext_{\bphi_i}(\bsx)$ and $\bsv_i := f_{\btheta_i}(\bsx) = \fcls_{\bpsi_i}(\bsz_i)$ for $i \in \{1, \dots, m\}$.  Let $\bsv_{i,j}(r) := f_{\bthetabe_{i,j}(r)}(\bsx)$. In order to use $\bsv_{i,j}(r)$ with $\bsv_i$ to get an ensemble, we should forward $\bsx$ through $f_{\bthetabe_{i,j}(r)}$, starting from the bottom layer. Instead, we  \emph{reuse} $\bsz_i$ to predict $\bsv_{i,j}(r)$ with a lightweight neural network. We call such a lightweight neural network a ``bridge network", since it allows us to move directly from $\bsv_i$ to $\bsv_{i,j}(r)$ in the function space, not through the actual parameter space. A bridge network is usually constructed with a \gls{cnn} whose inference cost is much lower than that of $f_{\btheta_i}$.

From the following, we introduce two types of bridge networks depending on the number of modes involved in the computation.
\cref{fig:flow_comparison} presents a schematic diagram that compares forward passes of ensembles with/without bridge networks.

\paragraph{Type I bridge networks}
A type I bridge network $\typeone_{i,j}$ takes a feature $\bsz_i$ from one mode and predicts $\bsv_{i,j}(r)$ as
\[
\bsv_{i,j}(r) \approx \tilde{\bsv}_{i,j}(r) = \typeone_{i,j}(\bsz_i).
\]
A type I bridge network can be constructed between any pair of connected modes $(\btheta_i, \btheta_j)$ and an ensembled prediction for specific mode $\btheta_i$ with its Bezier parameter $\bthetabe_{i,j}$ can be approximated as
\[
\frac{1}{2} \bigg( \bsv_i + \typeone_{i,j}(\bsz_i) \bigg),
\]

whose inference cost is nearly identical to that of $\bsv_i$ (nearly a single forward pass). One can also connect $\btheta_i$ with multiple modes $\{\btheta_{j_1}, \dots, \btheta_{j_k}\}$, learn bridge networks between $(i,j_1), \dots, (i, j_k)$, and construct an ensemble
\[
\frac{1}{1+k} \bigg(\bsv_i + \sum_{j=1}^k \typeone_{i, j_k}(\bsz_i)\bigg).
\label{eq:multi_type_one}
\]
Still, since the costs for $\typeone_{i,j_k}$s are much lower than $\bsv_i$, the inference cost does not increase significantly.

\paragraph{Type II bridge networks}
A type II bridge network between $(\btheta_i, \btheta_j)$ takes two features $(\bsz_i, \bsz_j)$ to predict $\bsv_{i,j}(r)$.
\[
\bsv_{i,j}(r) \approx \tilde{\bsv}_{i,j}(r) =  \typetwo_{i,j}(\bsz_i, \bsz_j).
\]
An ensembled prediction with the type II bridge network is then constructed as
\[
\frac{1}{3} \bigg( \bsv_i + \bsv_j + \typetwo_{i,j}(\bsz_i, \bsz_j)\bigg),
\]
where we construct an ensemble of three models with effectively two forward passes (for $\bsv_i$ and $\bsv_j$). Similar to the type I bridge networks, we may construct multiple bridges between modes and use them together for an ensemble.

\subsection{Learning bridge networks}

\begin{figure*}[t]
    \centering
    \begin{subfigure}[b]{0.98\textwidth}
        \includegraphics[width=\linewidth]{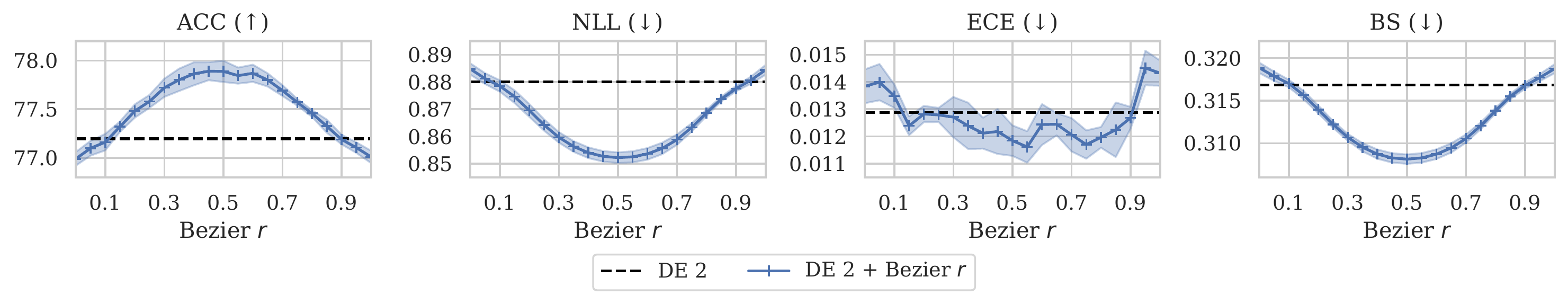}
        \label{fig:bezier_r_move}
        \vspace*{-4mm}
    \end{subfigure}
    \begin{subfigure}[b]{0.98\textwidth}
        \includegraphics[width=\linewidth]{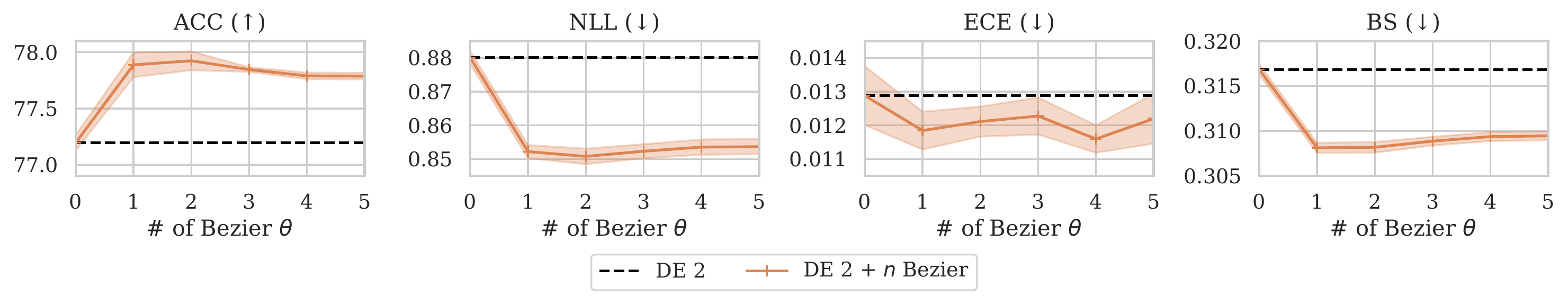}
        \label{fig:bezier_r_add}
        \vspace*{-6.5mm}
    \end{subfigure}
    \caption{Performance of an ensemble of two modes and a parameter from the Bezier curve connecting them, evaluated for ResNet-50 on ImageNet. Here, $r\in(0,1)$ denotes a position on the curve. (\textbf{Top}) Ensemble performances when one member from the Bezier curve $r$ is added to \gls{de}-2. (\textbf{Bottom}) Ensemble performances when members are sequentially added to \gls{de}-2 from a Bezier curve. For accuracy, the higher is better, and for NLL, ECE and BS, the lower is better.}
    \label{fig:bezier_curve}
    \vspace{-2.5mm}
\end{figure*}

\paragraph{Fixing a position $r$ on Bezier curves}
In the definition of the bridge networks above, we fixed the value $r$. In principle, we may parameterize the bridge networks to take $r$ as an additional input to predict $\bsv_{i,j}(r)$ for any $r \in [0, 1]$, but we found this to be ineffective due to the difficulty of learning all the outputs corresponding to arbitrary $r$ values. Moreover, as we empirically observed in \cref{fig:bezier_curve}, the ensemble with Bezier parameters is most effective with $r=0.5$, and adding additional parameters evaluated at different $r$ values does not significantly improve the performance. To this end, we fix $r=0.5$ and aim to learn bridge networks predicting $\bsv_{i,j}(0.5)$ throughout the paper. See \cref{sec:r_ablation} for more detailed inspection.

\paragraph{Training procedure}
Let $\{\btheta_1, \dots, \btheta_m\}$ be a set of parameters in an ensemble. Given a set of Bezier parameters $\{\bthetabe_{i,j}\}$ connecting them, we learn bridge networks (either type I or II) for each Bezier curve. The training procedure is straightforward. We minimize the Kullback-Leibler divergence between the actual output from the Bezier parameters and the prediction made from the bridge network. It makes the bridge network imitate the original function defined by the Bezier parameters in the same manner as a conventional knowledge distillation~\citep{hinton2015distilling}.

Further, we apply the mixup~\citep{zhang2018mixup} method to explore more diverse responses, preventing the bridge from learning to just copy the outputs of the base model.
Refer to~\cref{sec:training_procedure} for the detailed training procedure.

\section{Related Works}
\label{main:sec:related_works}

\paragraph{Mode connectivity}
The geometric properties of deep neural networks' loss surfaces have been studied, and one notable property is the mode connectivity~\citep{garipov2018loss,draxler2018essentially}; there exists a simple path between modes of a neural network on which the network retains low training error along that path.
From this, fast ensembling methods that collect ensemble members on the mode-connecting-paths have been proposed~\citep{huang2017snapshot,garipov2018loss}.
Extending this idea, \citet{izmailov2020subspace} approximated the posteriors of Bayesian neural nets via the low-loss subspace and used them for \gls{bma}.
\citet{wortsman2021learning} also presented a method for further improving performance by ensembling over the subspaces.

\paragraph{Efficient ensembling}
Despite the superior performance of \gls{de}~\citep{lakshminarayanan2017simple,ovadia2019trust}, it suffers from additional computation costs for both the training and the inference. There have been several works that reduce the computational burden in training by collecting ensemble members efficiently~\citep{huang2017snapshot,garipov2018loss,benton2021loss}, but they did not consider inference costs that arose from multiple forward passes.
On the other hand, there also exist inference-efficient ensembling methods by sharing parameters \citep{wen2019batchensemble,dusenberry2020efficient} or sharing representations \citep{lee2015m,siqueira2018ensemble,antoran2020depth,havasi2020training}. In particular, 
\citet{antoran2020depth} and \citet{havasi2020training} presented the methods to obtain an ensemble prediction by a single forward pass. Nevertheless, these methods do not scale well for complex large-scale datasets or require large network capacity.

\section{Experiments}
\label{main:sec:experiments}

In this section, we are going to answer the following three big questions:
\begin{enumerate}[leftmargin=*,itemsep=-0.7mm]
    \vspace{-3.5mm}
    \item Do bridge networks really learn to predict the outputs of a function from the Bezier curves?
    \item How much ensemble gain we obtain via bridge networks with lower computational complexity?
    \item How many bridge networks do we have to make in order to achieve certain ensemble performances?
    \vspace{-3.5mm}
\end{enumerate}
We answer them in \cref{sec:correspondence,sec:classification,sec:number_of_bridge} with empirical validation.

\begin{figure*}[t]
    \centering
    \includegraphics[width=0.97\textwidth]{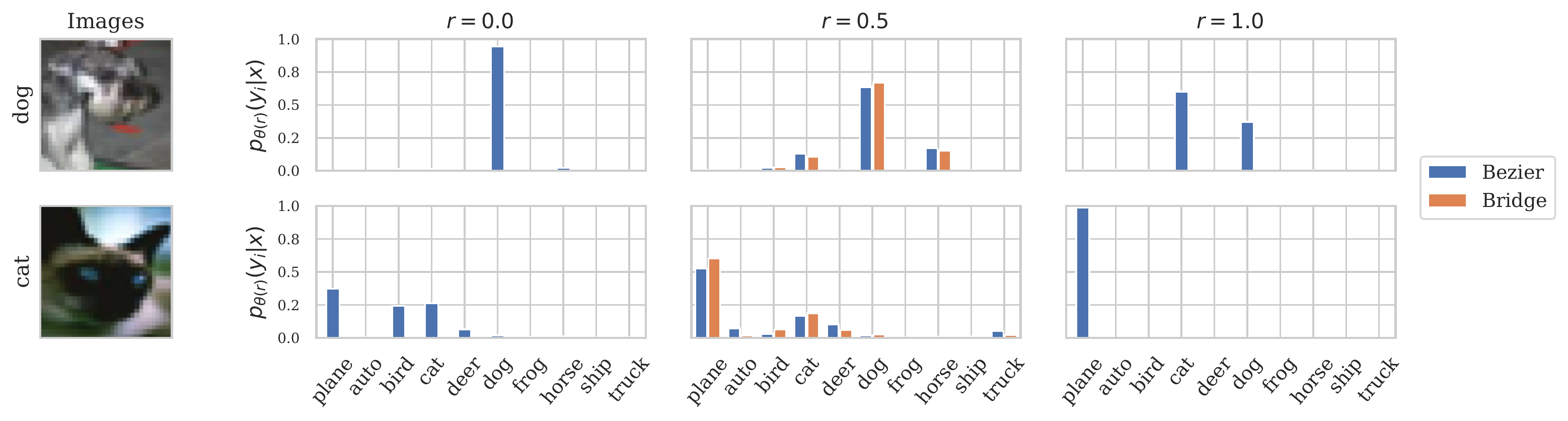}
    \vspace{-3mm}
    \caption{Bar plots in the third column show the class probability outputs of the bridge network (\textbf{orange}) and the base model with the Bezier parameters (\textbf{blue}) for given images displayed in the first column. We also depict the predicted outputs from the base model with $\btheta_1$ and $\btheta_2$ in the second and fourth columns, respectively. Additional results are in~\cref{sec:additional_examples}.}
    \label{fig:logit_regression}
    % \vspace{-3mm}
\end{figure*}

\begin{table*}[t]
    \centering
    \caption{$R^2$ scores and KL divergence values quantify how similar the following models to the target function defined with Bezier parameters $\bthetabe_{1,2}(0.5)$ are in output probabilities; `Match Type I/II Bridge', `Other Type I/II Bridge', and `Other Bezier'. Refer to~\cref{sec:correspondence} for a detailed description of each model. All values are measured on the test split of each dataset.}
    \vspace{-2mm}
    \resizebox{\textwidth}{!}{
        \begin{tabular}{lllllllll}
            \toprule
            Dataset              & \multicolumn{2}{c}{CIFAR-10}                & \multicolumn{2}{c}{CIFAR-100}               & \multicolumn{2}{c}{Tiny ImageNet}           & \multicolumn{2}{c}{ImageNet}                \\
                                   \cmidrule(lr){2-3}                            \cmidrule(lr){4-5}                            \cmidrule(lr){6-7}                            \cmidrule(lr){8-9}
            Model                & \umet{$R^2$}         & \dmet{KL}            & \umet{$R^2$}         & \dmet{KL}            & \umet{$R^2$}         & \dmet{KL}            & \umet{$R^2$}         & \dmet{KL}            \\
            \midrule
            Match type I bridge  & \tv[b]{0.916}{0.003} & \tv[b]{0.131}{0.005} & \tv[b]{0.805}{0.006} & \tv[b]{0.388}{0.011} & \tv[b]{0.751}{0.498} & \tv[b]{0.498}{0.015} & \tv[b]{0.906}{0.001} & \tv[b]{0.229}{0.002} \\
            Other type I bridge  & \tv[n]{0.908}{0.002} & \tv[n]{0.148}{0.004} & \tv[n]{0.780}{0.005} & \tv[n]{0.450}{0.010} & \tv[n]{0.719}{0.588} & \tv[n]{0.588}{0.045} & \tv[n]{0.894}{0.001} & \tv[n]{0.260}{0.004} \\
            \midrule
            Match type II bridge & \tv[b]{0.930}{0.002} & \tv[b]{0.108}{0.003} & \tv[b]{0.837}{0.002} & \tv[b]{0.318}{0.004} & \tv[b]{0.767}{0.004} & \tv[b]{0.459}{0.009} & \tv[b]{0.920}{0.001} & \tv[b]{0.191}{0.003} \\
            Other type II bridge & \tv[n]{0.911}{0.002} & \tv[n]{0.144}{0.003} & \tv[n]{0.794}{0.003} & \tv[n]{0.425}{0.006} & \tv[n]{0.720}{0.008} & \tv[n]{0.573}{0.011} & \tv[n]{0.898}{0.000} & \tv[n]{0.241}{0.002} \\
            \midrule
            Other Bezier         & \tv[n]{0.870}{0.004} & \tv[n]{0.229}{0.007} & \tv[n]{0.734}{0.002} & \tv[n]{0.586}{0.002} & \tv[n]{0.655}{0.701} & \tv[n]{0.701}{0.011} & \tv[n]{0.874}{0.002} & \tv[n]{0.323}{0.004} \\
            \bottomrule
        \end{tabular}
    }
    \label{tab:table_r_square}
    \vspace{-3mm}
\end{table*}

\subsection{Setup}

\paragraph{Datasets and networks}
We evaluate the proposed bridge networks on various image classification benchmarks, including CIFAR-10, CIFAR-100, Tiny ImageNet, and ImageNet datasets.
Throughout the experiments, we use the family of residual networks \citep[ResNet;][]{he2016deep} as a base model: ResNet-32$\times$2 for CIFAR-10, ResNet-32$\times$4 for CIFAR-100, ResNet-34 for Tiny ImageNet and ResNet-50 for ImageNet, where $\times$2 and $\times$4 denotes the multiplier of the number of channels for convolutional layers.
The base models for CIFAR datasets have fewer parameters than the (Tiny) ImageNet base models, which have different settings.
We construct bridge networks with \glspl{cnn} with a residual path whose inference costs are relatively low compared to those of ResNet base models.
For detailed training settings, including bridge network architectures or hyperparameter settings, please refer to \cref{sec:train_setting}.

By changing the channel size of the convolutional layers in the bridge network, we can balance the trade-off between performance gains with computational costs. We check this trade-off in \cref{sec:model_size_and_regression}. We refer to a bridge network with less than 10\% of floating-point operations (FLOPs) compared to the base model as Bridge\brgs (small bridge), and a bridge with more than 15\% as Bridge\brgm (medium bridge).

\vspace{-2mm}

\paragraph{Efficiency metrics}
We choose FLOPs and the number of parameters (\#Params) for efficiency evaluation as these metrics are commonly used to consider the efficiency~\citep{dehghani2021efficiency}.
Because FLOPs and \#Params of the base model are different for each dataset, we report the relative FLOPs and the relative \#Params with respect to the corresponding base model instead for better comparison.

\vspace{-2mm}

\paragraph{Uncertainty metrics}
As suggested by~\citet{ashukha2020pitfalls}, along with the classification accuracy (ACC), we report the calibrated versions of Negative Log-likelihood (NLL), Expected Calibration Error (ECE), and Brier Score (BS) as metrics for uncertainty evaluation.
We also measure the Deep Ensemble Equivalent (DEE) score proposed in~\citet{ashukha2020pitfalls}, which shows the relative performance for DE in terms of NLL and roughly be interpreted as \emph{effective number of models} for an ensemble. See~\cref{sec:metrics} for more details.

\vspace{-0.5mm}

\subsection{Correspondence between bridges and Beziers}
\label{sec:correspondence}

\cref{fig:logit_regression} visualizes the outputs of the bridge network $H_{1,2}^{(0.5)}$ which predicts the logits from $\bthetabe_{1,2}(0.5)$.
To be more specific, we visualize the predicted logits from $\btheta_1$, $\btheta_2$, $\bthetabe_{1,2}(0.5)$, and the bridge network $H_{1,2}^{(0.5)}$, for two test examples of CIFAR-10. Indeed, the bridge network predicts well the logits from the Bezier curve. \cref{sec:additional_examples} provides additional examples that further verify this.

To assess the quality of the prediction of bridge networks, we use a set of ensemble parameters $\{\btheta_1, \btheta_2, \dots, \btheta_m\}$ and Bezier curves between them. If the bridge network $H_{1,2}^{(0.5)}$ predicts $\bsv_{1,2}(0.5)$ well compared to the other baselines, we can confirm that there exists the correspondence between the bridge network and the Bezier curve. To this end, we measure the $R^2$ score and  Kullback-Leibler divergence (KL) which quantify how similar outputs of the following baselines to that of the target function $f_{\bthetabe_{1,2}(0.5)}$; (1) `Match type I/II bridge' denote the bridge network imitating the function of $\bthetabe_{1,2}(0.5)$, (2) `Other type I/II bridge' denote the bridge network imitating the function of $\bthetabe_{i,j}(0.5)$ for some $(i,j)\neq(1,2)$, and (3) `Other Bezier' denotes the base model with the parameters $\bthetabe_{i,j}(0.5)$ for some $(i,j)\neq(1,2)$.

\cref{tab:table_r_square} summarizes the results. Compared to the baselines (i.e., `Other type I/II bridge' and `Other Bezier'), the bridge networks produce more similar outputs to the target outputs. The $R^2$ values between the predictions and targets are significantly higher than those from the wrong targets, demonstrating that the bridge predictions indeed are approximating our target outputs of interest.

\begin{table*}[t]
    \newcommand{\metricrule}{\cmidrule(lr){2-3} \cmidrule(lr){4-7}}
    \newcommand{\modelrule}{\cmidrule(lr){1-7}}
    \centering
    \vspace{-1mm}
    \caption{
        Performance improvement of the ensemble by adding type I bridges to the single base ResNet model on Tiny ImageNet and ImageNet datasets.
        FLOPs, \#Params, and DEE metrics are measured with respect to the single base model.
        Bridge\brgs \ and Bridge\brgm \ denote the small and the medium versions of the bridge network based on their FLOPs.
    }
    \vspace{-3mm}
    \parbox{0.6\textwidth}{%
        \resizebox{\linewidth}{!}{%
            \begin{tabular}{lrrlllll}
                \textbf{Tiny ImageNet} \\
                \toprule
                Model                 & \dmet{FLOPs} & \dmet{\#Params} & \umet{ACC}          & \dmet{NLL}           & \dmet{ECE}           & \umet{DEE}           \\
                \midrule
                ResNet (\gls{de}-1)   & \relv{1.000} & \relv{1.000}    & \tv[n]{62.66}{0.23} & \tv[n]{1.683}{0.009} & \tv[n]{0.050}{0.004} & \tv[n]{1.000}{}      \\
                \metricrule
                \quad + 1 Bridge\brgs & \relv{1.050} & \relv{1.057}    & \tv[n]{64.58}{0.17} & \tv[n]{1.478}{0.006} & \tv[n]{0.025}{0.002} & \tv[n]{2.280}{0.086} \\
                \quad + 2 Bridge\brgs & \relv{1.099} & \relv{1.114}    & \tv[n]{65.37}{0.13} & \tv[n]{1.421}{0.004} & \tv[n]{0.018}{0.002} & \tv[n]{3.087}{0.118} \\
                \quad + 3 Bridge\brgs & \relv{1.149} & \relv{1.171}    & \tv[b]{65.82}{0.10} & \tv[b]{1.395}{0.003} & \tv[b]{0.015}{0.001} & \tv[b]{3.680}{0.133} \\
                \metricrule
                \quad + 1 Bridge\brgm & \relv{1.180} & \relv{1.206}    & \tv[n]{65.13}{0.12} & \tv[n]{1.446}{0.002} & \tv[n]{0.034}{0.002} & \tv[n]{2.709}{0.049} \\
                \quad + 2 Bridge\brgm & \relv{1.359} & \relv{1.412}    & \tv[n]{66.29}{0.06} & \tv[n]{1.388}{0.004} & \tv[n]{0.025}{0.001} & \tv[n]{3.845}{0.171} \\
                \quad + 3 Bridge\brgm & \relv{1.539} & \relv{1.618}    & \tv[b]{66.76}{0.09} & \tv[b]{1.362}{0.003} & \tv[b]{0.023}{0.001} & \tv[b]{4.708}{0.209} \\
                \modelrule
                \gls{de}-2            & \relv{2.000} & \relv{2.000}    & \tv[n]{65.54}{0.25} & \tv[n]{1.499}{0.007} & \tv[n]{0.029}{0.003} & \tv[n]{2.000}{}      \\
                \bottomrule \\[-1mm]

                \textbf{ImageNet} \\
                \toprule
                Model                 & \dmet{FLOPs} & \dmet{\#Params} & \umet{ACC}          & \dmet{NLL}           & \dmet{ECE}           & \umet{DEE}           \\
                \midrule
                ResNet (\gls{de}-1)   & \relv{1.000} & \relv{1.000}    & \tv[n]{75.85}{0.06} & \tv[n]{0.936}{0.003} & \tv[n]{0.019}{0.001} & \tv[n]{1.000}{}      \\
                \metricrule
                \quad + 1 Bridge\brgs & \relv{1.061} & \relv{1.071}    & \tv[n]{76.46}{0.06} & \tv[n]{0.914}{0.000} & \tv[n]{0.012}{0.001} & \tv[n]{1.418}{0.034} \\
                \quad + 2 Bridge\brgs & \relv{1.123} & \relv{1.141}    & \tv[n]{76.60}{0.06} & \tv[n]{0.907}{0.000} & \tv[n]{0.012}{0.001} & \tv[n]{1.537}{0.026} \\
                \quad + 3 Bridge\brgs & \relv{1.184} & \relv{1.212}    & \tv[b]{76.69}{0.04} & \tv[b]{0.905}{0.000} & \tv[b]{0.011}{0.001} & \tv[b]{1.584}{0.021} \\
                \metricrule
                \quad + 1 Bridge\brgm & \relv{1.194} & \relv{1.222}    & \tv[n]{77.03}{0.07} & \tv[n]{0.889}{0.001} & \tv[b]{0.013}{0.000} & \tv[n]{1.881}{0.022} \\
                \quad + 2 Bridge\brgm & \relv{1.389} & \relv{1.444}    & \tv[n]{77.37}{0.07} & \tv[n]{0.876}{0.001} & \tv[b]{0.013}{0.001} & \tv[n]{2.341}{0.076} \\
                \quad + 3 Bridge\brgm & \relv{1.583} & \relv{1.665}    & \tv[b]{77.48}{0.03} & \tv[b]{0.870}{0.000} & \tv[b]{0.013}{0.000} & \tv[b]{2.618}{0.062} \\
                \modelrule
                \gls{de}-2            & \relv{2.000} & \relv{2.000}    & \tv[n]{77.12}{0.04} & \tv[n]{0.883}{0.001} & \tv[n]{0.012}{0.001} & \tv[n]{2.000}{}      \\
                \bottomrule
            \end{tabular}
        }
    }
    \label{tab:type1}
    \parbox{0.37\textwidth}{%
        \vspace{2mm}
        \resizebox{\linewidth}{!}{%
            \includegraphics{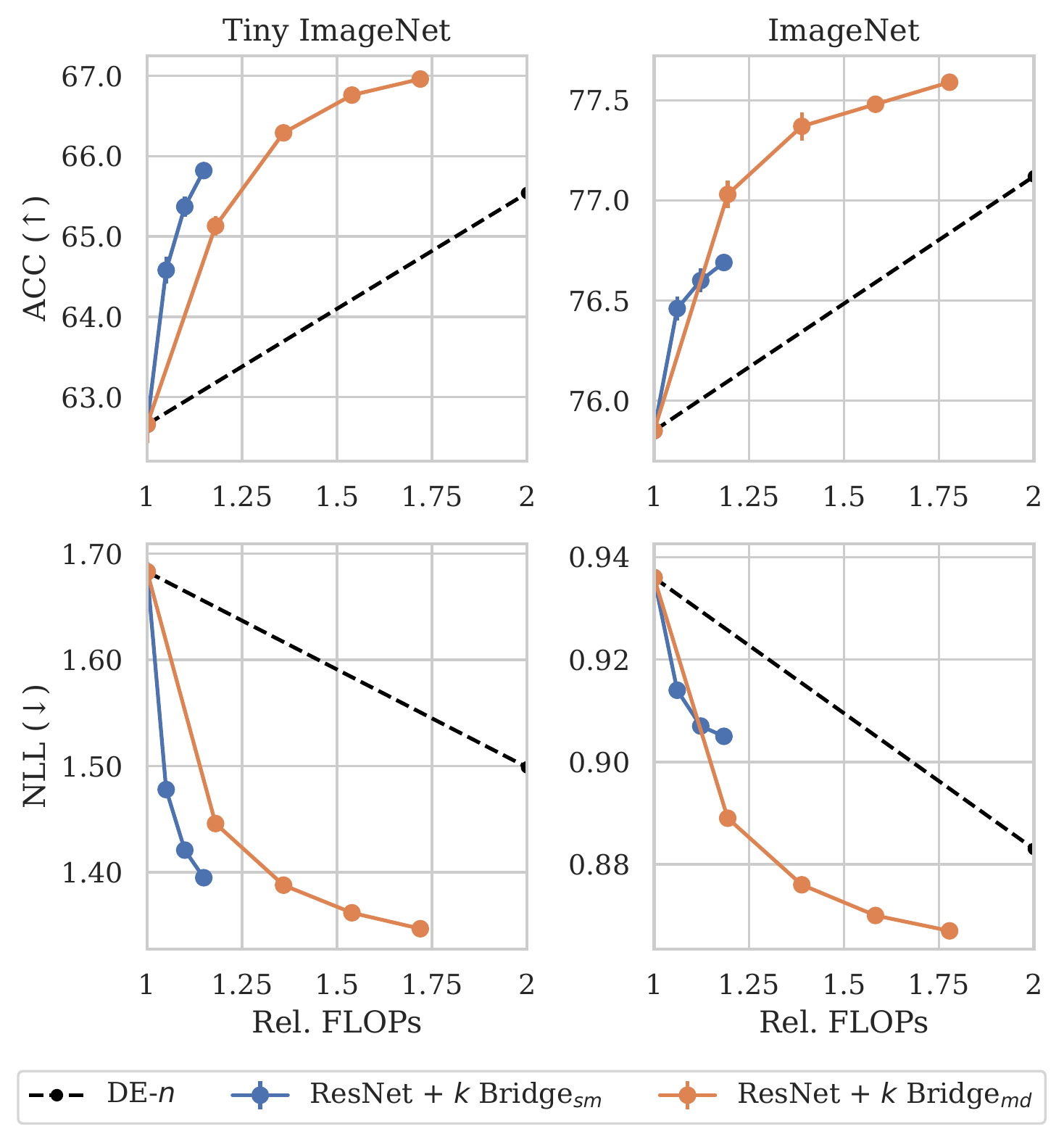}
        }
    }
    \vspace{-2mm}
    \captionof{figure}{
        The cost-performance plots of type I bridges compared to \gls{de} on Tiny ImageNet and ImageNet datasets.
        The x-axis denotes the relative FLOPs quantifying the inference cost of the model compared to a single base model, and the y-axis shows the corresponding predictive performance.
        On the basis of \gls{de} (black dashed line), the upper left position is preferable in ACC, and the lower left position is preferable in NLL.
    }
    \label{fig:type1}
    \vspace{-3mm}
\end{table*}

\subsection{Classification with bridge networks}
\label{sec:classification}

\subsubsection{Type I bridge networks}
\label{sec:type1_experiment}

\paragraph{Single model performances with type I bridge networks}
In situations where multiple forward passes are not allowed for inference, we can approximate an ensemble of a single base model and the ones from Bezier curves with type I bridge networks.
The results are shown in \cref{tab:type1}. The results show that for \gls{sgd} trained single ResNet model, an ensemble with type I bridge networks improves the performance both in terms of accuracy and uncertainty estimation.
Only adding one small type I bridge with $5 \sim 6\%$ FLOPs to the base model (ResNet + 1 Bridge\brgs) dramatically improves the accuracy and DEE on both datasets.
Furthermore, using more FLOPs with a medium type I bridge (ResNet + 1 Bridge\brgm) gives better performance gains.

\vspace{-2mm}

\paragraph{Using multiple type I bridge networks}
As type I bridge network requires features from only one mode of each curve for inference, we can use multiple type I bridge networks for a single base model without significantly increasing inference cost, as we mentioned at \cref{eq:multi_type_one}.
\cref{tab:type1} reports the performance gain of a single base model with an increasing number of type I bridges.
Each bridge approximates the models on different Bezier curves between a single mode and others (i.e., Bezier curves between modes A-B, A-C, and so on where A, B, and C are different modes.), not the models on a single Bezier curve. Adding more bridge networks introduces more diverse outputs to the ensembles.
One can see that the performance continuously improves as the number of bridges increases, with low additional inference costs.
\cref{fig:type1} shows how much type I bridge networks efficiently increase the performances proportional to FLOPs.

\vspace{-1mm}

\subsubsection{Type II bridge networks}

\begin{table*}[t]
    \newcommand{\metricrule}{\cmidrule(lr){2-3} \cmidrule(lr){4-7}}
    \newcommand{\modelrule}{\cmidrule(lr){1-7}}
    \centering
    \vspace{-1mm}
    \caption{
        Performance improvement of the ensemble by adding type II bridges as members to existing DE ensembles on Tiny ImageNet and ImageNet datasets.
        FLOPs, \#Params, and DEE metrics are measured with respect to corresponding DEs.
        Type II bridges consistently improve the accuracy and uncertainty metrics of the ensemble before saturation.
        Bridge\brgs \ and Bridge\brgm \ denote the small and the medium versions of the bridge network based on their FLOPs.
    }
    \vspace{-3mm}
    \def\arraystretch{0.95}
    \parbox{0.6\textwidth}{%
        \resizebox{\linewidth}{!}{%
            \begin{tabular}{lrrlllll}
                \textbf{Tiny ImageNet} \\
                \toprule
                Model                 & \dmet{FLOPs} & \dmet{\#Params} & \umet{ACC}          & \dmet{NLL}           & \dmet{ECE}           & \umet{DEE}           \\
                \midrule
                \gls{de}-4            & \relv{4.000} & \relv{4.000}    & \tv[n]{67.50}{0.11} & \tv[n]{1.381}{0.004} & \tv[n]{0.018}{0.001} & \tv[n]{ 4.000}{}      \\
                \metricrule
                \quad + 1 Bridge\brgs & \relv{4.058} & \relv{4.067}    & \tv[n]{67.86}{0.05} & \tv[n]{1.334}{0.003} & \tv[n]{0.017}{0.002} & \tv[n]{ 6.051}{0.181} \\
                \quad + 2 Bridge\brgs & \relv{4.117} & \relv{4.135}    & \tv[n]{68.12}{0.09} & \tv[n]{1.311}{0.005} & \tv[n]{0.015}{0.001} & \tv[n]{ 8.174}{0.465} \\
                \quad + 4 Bridge\brgs & \relv{4.234} & \relv{4.269}    & \tv[n]{68.47}{0.14} & \tv[n]{1.288}{0.004} & \tv[n]{0.015}{0.001} & \tv[n]{10.340}{0.773} \\
                \quad + 6 Bridge\brgs & \relv{4.351} & \relv{4.404}    & \tv[b]{68.51}{0.10} & \tv[b]{1.278}{0.003} & \tv[b]{0.014}{0.001} & \tv[b]{11.268}{0.871} \\
                \metricrule
                \quad + 1 Bridge\brgm & \relv{4.198} & \relv{4.226}    & \tv[n]{68.00}{0.11} & \tv[n]{1.333}{0.003} & \tv[b]{0.019}{0.001} & \tv[n]{ 6.183}{0.120} \\
                \quad + 2 Bridge\brgm & \relv{4.395} & \relv{4.453}    & \tv[n]{68.33}{0.08} & \tv[n]{1.308}{0.003} & \tv[b]{0.019}{0.001} & \tv[n]{ 8.489}{0.481} \\
                \quad + 4 Bridge\brgm & \relv{4.791} & \relv{4.906}    & \tv[n]{68.61}{0.05} & \tv[n]{1.281}{0.004} & \tv[n]{0.021}{0.003} & \tv[n]{10.897}{0.800} \\
                \quad + 6 Bridge\brgm & \relv{5.186} & \relv{5.359}    & \tv[b]{68.80}{0.09} & \tv[b]{1.269}{0.003} & \tv[n]{0.021}{0.001} & \tv[b]{12.110}{1.083} \\
                \modelrule
                \gls{de}-5            & \relv{5.000} & \relv{5.000}    & \tv[n]{67.90}{0.14} & \tv[n]{1.354}{0.003} & \tv[n]{0.019}{0.001} & \tv[n]{ 5.000}{}      \\
                \bottomrule \\[-1mm]

                \textbf{ImageNet} \\
                \toprule
                Model                 & \dmet{FLOPs} & \dmet{\#Params} & \umet{ACC}          & \dmet{NLL}           & \dmet{ECE}           & \umet{DEE}           \\
                \midrule
                \gls{de}-4            & \relv{4.000} & \relv{4.000}    & \tv[n]{77.87}{0.04} & \tv[n]{0.851}{0.001} & \tv[n]{0.012}{0.001} & \tv[n]{ 4.000}{}      \\
                \metricrule
                \quad + 1 Bridge\brgs & \relv{4.086} & \relv{4.088}    & \tv[n]{77.93}{0.02} & \tv[n]{0.847}{0.000} & \tv[n]{0.012}{0.001} & \tv[n]{ 4.580}{0.052} \\
                \quad + 2 Bridge\brgs & \relv{4.172} & \relv{4.176}    & \tv[n]{78.00}{0.04} & \tv[b]{0.846}{0.000} & \tv[b]{0.011}{0.000} & \tv[n]{ 4.739}{0.052} \\
                \quad + 4 Bridge\brgs & \relv{4.343} & \relv{4.351}    & \tv[n]{78.10}{0.03} & \tv[b]{0.846}{0.000} & \tv[b]{0.011}{0.001} & \tv[b]{ 4.768}{0.041} \\
                \quad + 6 Bridge\brgs & \relv{4.515} & \relv{4.527}    & \tv[b]{78.12}{0.05} & \tv[b]{0.846}{0.001} & \tv[b]{0.011}{0.001} & \tv[n]{ 4.659}{0.037} \\
                \metricrule
                \quad + 1 Bridge\brgm & \relv{4.243} & \relv{4.256}    & \tv[n]{78.14}{0.03} & \tv[n]{0.839}{0.000} & \tv[b]{0.011}{0.001} & \tv[n]{ 6.123}{0.121} \\
                \quad + 2 Bridge\brgm & \relv{4.487} & \relv{4.512}    & \tv[n]{78.30}{0.05} & \tv[n]{0.833}{0.000} & \tv[n]{0.012}{0.001} & \tv[n]{ 8.068}{0.144} \\
                \quad + 4 Bridge\brgm & \relv{4.973} & \relv{5.024}    & \tv[n]{78.46}{0.04} & \tv[n]{0.828}{0.000} & \tv[n]{0.012}{0.000} & \tv[n]{ 9.951}{0.163} \\
                \quad + 6 Bridge\brgm & \relv{5.460} & \relv{5.536}    & \tv[b]{78.56}{0.09} & \tv[b]{0.825}{0.000} & \tv[n]{0.012}{0.001} & \tv[b]{10.760}{0.202} \\
                \modelrule
                \gls{de}-5            & \relv{5.000} & \relv{5.000}    & \tv[n]{78.03}{0.03} & \tv[n]{0.844}{0.001} & \tv[n]{0.012}{0.001} & \tv[n]{ 5.000}{}      \\
                \bottomrule
            \end{tabular}
        }
    }
    \label{tab:type2}
    \parbox{0.37\textwidth}{%
        \vspace{2mm}
        \resizebox{\linewidth}{!}{%
            \includegraphics{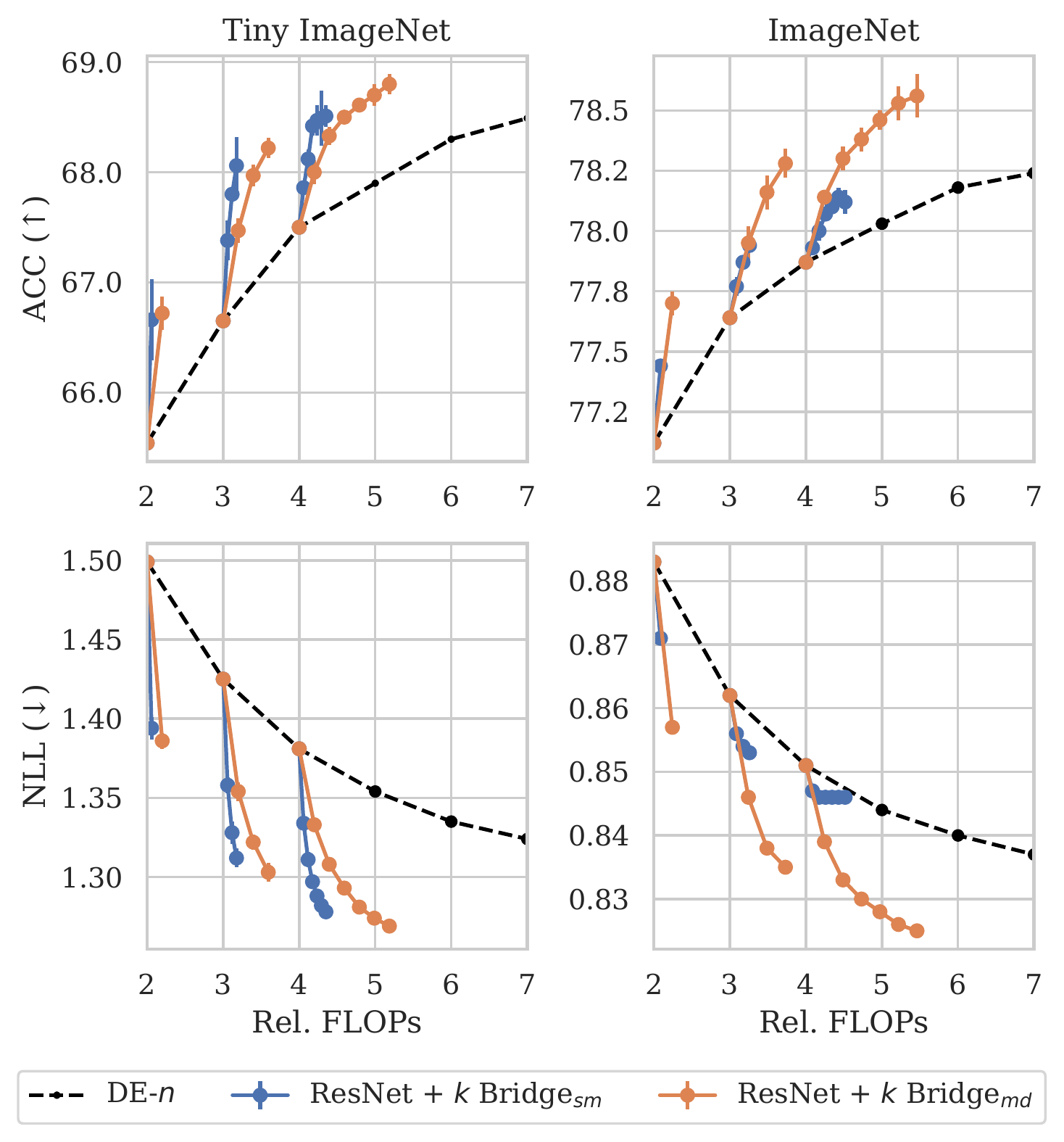}
        }
    }
    \vspace{-2mm}
    \captionof{figure}{
        The cost-performance plots of type II bridges compared to \gls{de} on Tiny ImageNet and ImageNet datasets.
        The x-axis denotes the relative FLOPs quantifying the inference cost of the model compared to \gls{de} basis from \gls{de}-2 to \gls{de}-7, and the y-axis shows the corresponding predictive performance.
        On the basis of \gls{de} (black dashed line), the upper left position is preferable in ACC, and the lower left position is preferable in NLL.
    }
    \label{fig:type2}
    \vspace{-3.5mm}
\end{table*}

\paragraph{Performance}
\cref{tab:type2} summarizes the classification results comparing \gls{de}, \gls{de} with Bezier curves, and \gls{de} with type II bridge networks. For more experimental results including other datasets, please refer to \cref{sec:full_results}. From \cref{tab:type2}, one can see that with only a slight increase in the computational costs, the ensembles with bridge networks achieve almost DEE 2.051 ensemble gain for \gls{de}-4 case on Tiny ImageNet dataset. This gain is not specific only for \gls{de}-4; the ensembles with type II bridge networks consistently improved predictive accuracy and uncertainty calibration with a small increase in the inference costs. \cref{fig:type2} shows how much our type II bridge network achieves high performance in the perspective of relative FLOPs.

\vspace{-2mm}

\paragraph{Computational cost}

We report FLOPs for inference on \cref{tab:type2} to indicate how much relative computational costs are required for the competing models. \cref{fig:type2} summarizes the tradeoff between FLOPs and performance in various metrics. As one can see from these results, our bridge networks could achieve a remarkable gain in performance, so for some cases, adding bridge ensembles achieved performance gains larger than those might be achieved by adding entire ensemble members. For instance, in Tiny ImageNet experiments, \gls{de}-4 + 2 bridges were better than \gls{de}-5.
Please refer to \cref{sec:full_results} for the full results including various \gls{de} sizes and other datasets.

\vspace{-2mm}

\subsection{How many type II bridges are required?}
\label{sec:number_of_bridge}
\vspace{-1mm}

For an ensemble of $m$ parameters, the number of pairs that can be connected by Bezier curves is $\binom{m}{2}$, which grows quadratically with $m$. In the previous experiment, we constructed Bezier curves and bridges for all possible pairs (which explains the large inference costs for Bezier ensembles), but in practice, we found that it is not necessary to use bridge networks for all of those pairs. As an example, we compare the performance of \gls{de}-4 + bridge ensembles with an increasing number of bridges on Tiny ImageNet and ImageNet datasets. The results are summarized in \cref{tab:type2}. Just one small bridge dramatically increases the performance, and the performance gain gradually saturates as we add more bridges. Notably, only one bridge shows similar or better performance than \gls{de}-5.

\begin{table*}[t]
    \newcommand{\metricrule}{\cmidrule(lr){2-3} \cmidrule(lr){4-7}}
    \newcommand{\modelrule}{\cmidrule(lr){1-7}}
    \newcommand{\methodrule}{\cmidrule(lr){1-1} \cmidrule(lr){2-3} \cmidrule(lr){4-7}}
    \centering
    \vspace{-1mm}
    \caption{
        Comparison of performance improvement of the efficient methods on Tiny ImageNet dataset.
        FLOPs, \#Params, and DEE metrics are measured with respect to the single ResNet-34.
    }
    \vspace{-3mm}
    \parbox{0.60\textwidth}{%
        \resizebox{\linewidth}{!}{%
            \begin{tabular}{lrrlllll}
                \textbf{Tiny ImageNet} \\
                \toprule
                Model                              & \dmet{FLOPs} & \dmet{\#Params} & \umet{ACC}          & \dmet{NLL}           & \dmet{ECE}           & \umet{DEE}           \\
                \midrule
                ResNet (\gls{de}-1)                & \relv{1.000} & \relv{1.000}    & \tv[n]{62.66}{0.23} & \tv[n]{1.683}{0.009} & \tv[n]{0.050}{0.004} & \tv[n]{1.000}{}      \\
                \metricrule
                \quad + 1 Bridge\brgs              & \relv{1.050} & \relv{1.057}    & \tv[n]{64.58}{0.17} & \tv[n]{1.478}{0.006} & \tv[n]{0.025}{0.002} & \tv[n]{2.280}{0.086} \\
                \quad + 2 Bridge\brgs              & \relv{1.099} & \relv{1.114}    & \tv[n]{65.37}{0.13} & \tv[n]{1.421}{0.004} & \tv[n]{0.018}{0.002} & \tv[n]{3.087}{0.118} \\
                \quad + 3 Bridge\brgs              & \relv{1.149} & \relv{1.171}    & \tv[n]{65.82}{0.10} & \tv[n]{1.395}{0.003} & \tv[n]{0.015}{0.001} & \tv[n]{3.680}{0.133} \\
                \modelrule
                \gls{de}-2                         & \relv{2.000} & \relv{2.000}    & \tv[n]{65.54}{0.25} & \tv[n]{1.499}{0.007} & \tv[n]{0.029}{0.003} & \tv[n]{2.000}{}      \\
                \gls{de}-3                         & \relv{3.000} & \relv{3.000}    & \tv[n]{66.65}{0.18} & \tv[n]{1.425}{0.005} & \tv[n]{0.024}{0.002} & \tv[n]{3.000}{}      \\
                \methodrule
                BE-2                               & \relv{2.000} & \relv{1.001}    & \tv[n]{61.44}{0.16} & \tv[n]{1.651}{0.006} & \tv[n]{0.014}{0.003} & \tv[n]{1.169}{0.079} \\
                BE-3                               & \relv{3.000} & \relv{1.002}    & \tv[n]{62.33}{0.24} & \tv[n]{1.603}{0.014} & \tv[n]{0.015}{0.003} & \tv[n]{1.433}{0.070} \\
                BE-4                               & \relv{4.000} & \relv{1.003}    & \tv[n]{62.42}{0.58} & \tv[n]{1.607}{0.020} & \tv[n]{0.015}{0.003} & \tv[n]{1.410}{0.098} \\
                \methodrule
                MIMO \small{($M = 2$)}             & \relv{2.000} & \relv{3.994}    & \tv[n]{63.47}{0.76} & \tv[n]{1.691}{0.013} & \tv[n]{0.065}{0.000} & \tv[n]{0.983}{0.053} \\
                \methodrule
                ResNet \small{(SWA)}               & \relv{1.000} & \relv{1.000}    & \tv[n]{64.03}{0.21} & \tv[n]{1.519}{0.010} & \tv[n]{0.030}{0.002} & \tv[n]{1.888}{0.074} \\
                \metricrule
                \quad + 1 Bridge\brgs              & \relv{1.050} & \relv{1.057}    & \tv[n]{65.26}{0.08} & \tv[n]{1.435}{0.004} & \tv[n]{0.031}{0.001} & \tv[n]{2.865}{0.087} \\
                \quad + 2 Bridge\brgs              & \relv{1.099} & \relv{1.114}    & \tv[n]{65.77}{0.10} & \tv[n]{1.403}{0.003} & \tv[n]{0.028}{0.002} & \tv[n]{3.511}{0.162} \\
                \quad + 3 Bridge\brgs              & \relv{1.149} & \relv{1.171}    & \tv[n]{65.96}{0.09} & \tv[n]{1.387}{0.001} & \tv[b]{0.026}{0.002} & \tv[n]{3.873}{0.160} \\
                \methodrule
                ResNet \tiny{(KD from \gls{de}-2)} & \relv{1.000} & \relv{1.000}    & \tv[n]{64.71}{0.23} & \tv[n]{1.629}{0.005} & \tv[n]{0.066}{0.002} & \tv[n]{1.290}{0.038} \\
                ResNet \tiny{(KD from \gls{de}-3)} & \relv{1.000} & \relv{1.000}    & \tv[n]{65.09}{0.25} & \tv[n]{1.622}{0.002} & \tv[n]{0.068}{0.003} & \tv[n]{1.331}{0.035} \\
                \bottomrule
            \end{tabular}
        }
    }
    \label{tab:comparison}
    \parbox{0.35\textwidth}{%
        \vspace{2mm}
        \resizebox{\linewidth}{!}{%
            \includegraphics{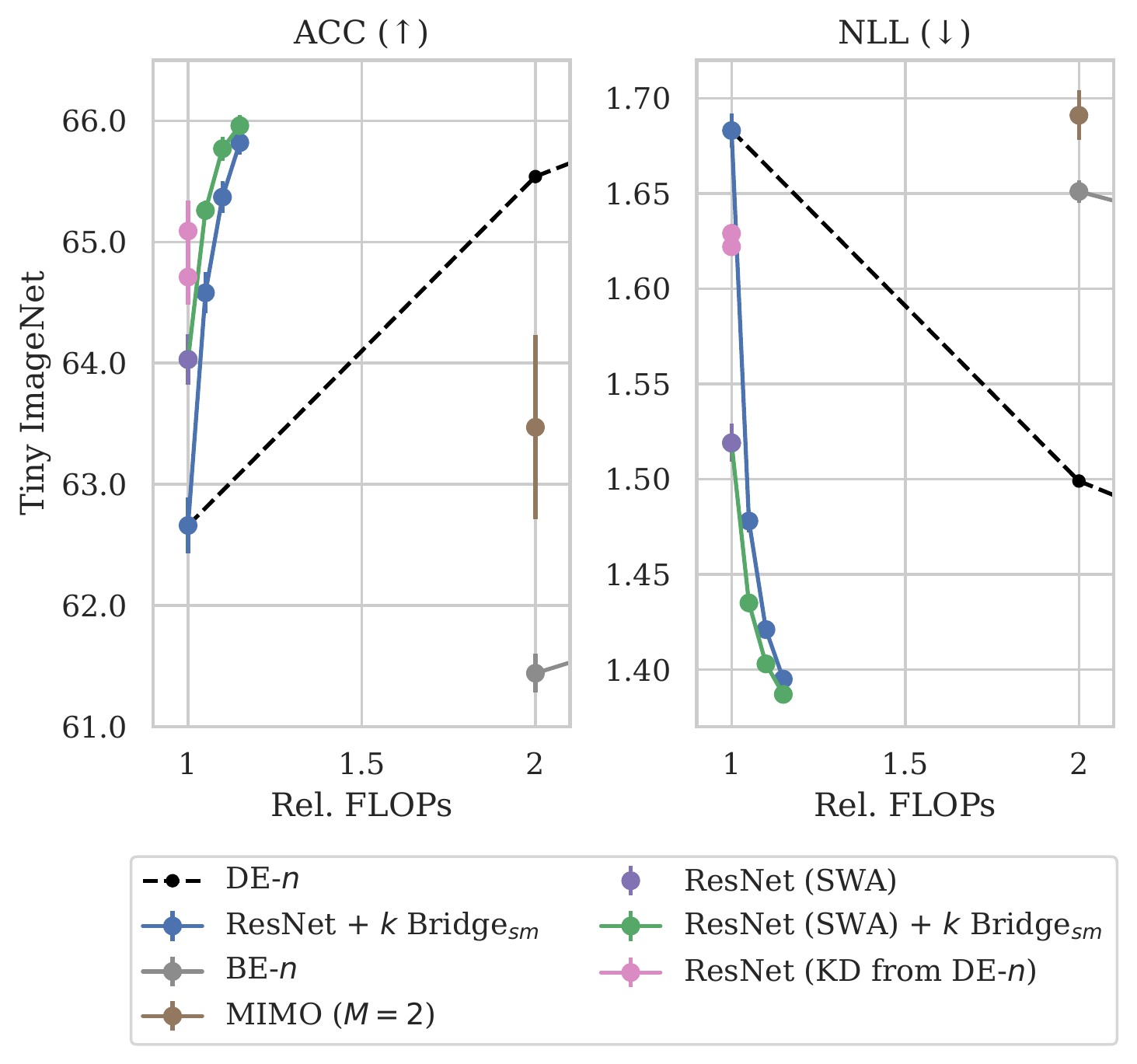}
        }
    }
    \vspace{-2mm}
    \captionof{figure}{
        The cost-performance plots of efficient methods compared to \gls{de} of ResNet-34 on Tiny ImageNet dataset.
        The x-axis denotes the relative FLOPs quantifying the inference cost of the model compared to a single ResNet-34, and the y-axis shows the corresponding predictive performance.
        On the basis of \gls{de} (black dashed line), the upper left position is preferable in ACC, and the lower left position is preferable in NLL.
    }
    \label{fig:comparison}
    \vspace{-4mm}
\end{table*}

\vspace{-1.9mm}

\subsection{Comparison with other efficient ensemble methods}
\label{sec:comparison}

To show the efficiency of our model, we compared it to methods that efficiently use an ensemble. For comparison, we choose BatchEnsemble~\citep[BE;][]{wen2019batchensemble}, multi-input multi-output network~\citep[MIMO;][]{havasi2020training}, a single model trained with stochastic weight averaging~\citep[SWA;][]{izmailov2018averaging}, and a single model that knowledge-distilled from \gls{de}-$n$ teacher models~\citep{hinton2015distilling} on Tiny ImageNet dataset.

\cref{tab:comparison} and \cref{fig:comparison} summarize the results. 
BatchEnsemble is efficient in terms of the number of parameters, but it still uses the same FLOPs as \gls{de}. In addition, the method is not scalable, so it does not work well for relatively difficult datasets such as Tiny ImageNet.

MIMO operates effectively when the capacity of the model is sufficient because several subnetworks are independently learned in a single model and the ensemble effect is expected with only one forward pass. However, in difficult datasets, it is hard to have sufficient network capacity, and it is not easy to train a model because additional training techniques such as batch repetition are required. Despite using ResNet-34$\times$2, which increased the width by twice as much as the baseline ResNet, the performance did not improve as expected.

SWA and knowledge distillation give considerable performance gain to a single model. However, by adding a bridge network of about 5\% FLOPs, it outperforms SWA in terms of both ACC and NLL.
And knowledge-distilled single models show high accuracies, but the advantage for calibration that can be obtained from the ensemble is not large.
As an extension of our method, a model that is trained in the form of a single model can be used as the base model of bridge networks to give a greater performance gain.
The results show that the type I bridges using an SWA-trained single model as a base model (ResNet (SWA) + Bridge\brgs) further improve the performance.
We empirically validate this in \cref{sec:swa_bridge}.
\section{Discussion}
\label{main:sec:discussion}

\paragraph{Difference between type I and type II bridges}
Both types I and II bridges approximate the model located at the midpoint of the Bezier curve with a small inference cost. While the type II bridge requires both outputs of models located at two endpoints of the Bezier curve, the type I bridge only requires one. Since we provide more information about the Bezier subspace to the type II bridge, it will approximate the midpoint of the Bezier curve more accurately than the type I bridge. Indeed, the experimental results presented in \cref{sec:correspondence} shows that the type II bridge produces more similar output to the target model (i.e., the midpoint of the Bezier curve) than the type I bridge. Nevertheless, it is worth investigating the type I bridge since its strength lies in the possibility of scaling up a single model. To be specific, \cref{sec:type1_experiment} shows that type I bridges can progressively enhance the single model to be stronger than \gls{de}-{1,2,3} with a relatively lower cost.

\vspace{-2.5mm}

\paragraph{Which type of bridge network should I use?}
Although type II bridges usually perform better than type I (due to the rich information that came from two endpoints of the Beizer curve), they presume the feasibility of multiple base networks during inference. Thus, in practice, it is encouraged to use type II bridges only if we are allowed to forward multiple base networks (if not, we should use type I bridges). If the available FLOPs and memory are between \gls{de}-1 and \gls{de}-2, only type I bridges can be used, because they can work with a single base model (one endpoint). If you have more FLOPs and more memory available than the \gls{de}-2, you can use type II bridges. Although type II bridges require base models for both endpoints, they can make more accurate predictions and show higher ensemble gains than type I bridges. This point can be confirmed more clearly by looking at the x-axis (relative FLOPs) of \cref{fig:type1} and \cref{fig:type2} of the paper, which shows the relative performance compared to the relative FLOPs of type I and type II bridges, and the performance improvement accordingly. In summary, type I bridges can be effectively used in a situation where available relative FLOPs are about $\times 1 \sim \times 2$, and type II bridges can be effectively used with type I bridges in a situation where available relative FLOPs are more than $\times 2$. We inspect the situation in which type I and type II bridges are mixed in \cref{sec:type_mix}.

\vspace{-1mm}

\section{Conclusion}
\label{main:sec:conclusion}

\vspace{-0.5mm}

In this paper, we proposed a novel framework for efficient ensembling that reduces inference costs of ensembles with a lightweight network called bridge networks. Bridge networks predict the neural network outputs corresponding to the parameters obtained from the Bezier curves connecting two ensemble parameters without actual forward passes through the network. Instead, they reuse features and outputs computed from the ensemble members and directly predict the outputs corresponding to Bezier parameters in function spaces. Using various image classification benchmarks, we demonstrate that we can train such bridge networks with simple \glspl{cnn} with minimal inference costs, and bridge-augmented ensembles could achieve significant gain in terms of accuracy and uncertainty calibration.

%%%%%%%%%%%%%%%%%%%%%%%%%%%%%%%%%%%%%%%%%%%%%%%%%%%%%%%%%%%%%%%%%%%%%%%%%%%%%%%
\section*{Acknowledgements}
% AI대학원/Fairness과제/삼성반도체과제
This work was partly supported by Institute of Information \& communications Technology Planning \& Evaluation (IITP) grant funded by the Korea government (MSIT) (No.2022-0-00184, Development and Study of AI Technologies to Inexpensively Conform to Evolving Policy on Ethics, and No.2019-0-00075, Artificial Intelligence Graduate School Program (KAIST)) and Samsung Electronics Co., Ltd. (IO201214-08176-01).
% TRC Research Cloud
We acknowledge the support of Google's TPU Research Cloud (TRC), which provided us with access to TPU v2 and TPU v3 accelerators for this work.
%%%%%%%%%%%%%%%%%%%%%%%%%%%%%%%%%%%%%%%%%%%%%%%%%%%%%%%%%%%%%%%%%%%%%%%%%%%%%%%

\bibliography{references}
\bibliographystyle{icml2023}

%%%%%%%%%%%%%%%%%%%%%%%%%%%%%%%%%%%%%%%%%%%%%%%%%%%%%%%%%%%%%%%%%%%%%%%%%%%%%%%
%%%%%%%%%%%%%%%%%%%%%%%%%%%%%%%%%%%%%%%%%%%%%%%%%%%%%%%%%%%%%%%%%%%%%%%%%%%%%%%
% APPENDIX
%%%%%%%%%%%%%%%%%%%%%%%%%%%%%%%%%%%%%%%%%%%%%%%%%%%%%%%%%%%%%%%%%%%%%%%%%%%%%%%
%%%%%%%%%%%%%%%%%%%%%%%%%%%%%%%%%%%%%%%%%%%%%%%%%%%%%%%%%%%%%%%%%%%%%%%%%%%%%%%
\newpage
\appendix
\onecolumn
\section{Experimental Details}
\label{sec:details}

We release the code used in the experiments on GitHub\footnote{\href{https://github.com/yuneg11/Bridge-Network}{https://github.com/yuneg11/Bridge-Network}}.

%%%%%%%%%%%%%%%%%%%%%%%%%%%%%%%%%%%%%%%%%%%%%%%%%%

\subsection{Training procedure}
\label{sec:training_procedure}

\vspace{-5mm}

\parbox{\textwidth}{
    \begin{minipage}{0.49\linewidth}
        \begin{algorithm}[H]
            \caption{Training type I bridge networks}
            \label{alg:bridge_training_type1}
            \begin{algorithmic}
                \REQUIRE{Training dataset $\calD$, a pair of parameters $(\btheta_1, \btheta_2)$ and corresponding Bezier parameter $\bthetabe_{1,2}$,
                 a bridge network $\typeone_{1,2}$ (type I) with parameters $\bomega$, learning rate $\eta$, a mixup coefficient $\alpha$.}
                \vspace{2mm}
                
                \STATE Initialize $\bomega$.
                \WHILE{not converged}
                    \STATE Sample a mini-batch $\calB \sim \calD$.
                    \FOR{$i=1,\dots,|\calB|$}
                        \STATE Take the input $\bsx_i$ from $\calB$.
                        \STATE $\tilde{\bsx_i} \gets \text{mixup}(\bsx_i, \alpha)$
                        \STATE $\bsz_1 \gets \fext_{\bphi_1}(\tilde{\bsx_i})$, $\bsv_1 \gets \fcls_{\bpsi_1}(\bsz_1)$.
                        \STATE $\bsv_{1,2}(r) \gets f_{\bthetabe_{1,2}(0.5)}(\tilde{\bsx_i})$.
                        \STATE $\tilde{\bsv}_{1,2}(r) \gets h_{1,2}^{(0.5)}(\bsz_1 ; \bomega)$.
                        \STATE $\ell_i \gets D_\text{KL}(\bsv_{1,2}(0.5) || \tilde\bsv_{1,2}(0.5))$
                    \ENDFOR
                    \STATE $\bomega\gets \bomega - \eta \nabla_\bomega \frac{1}{|\calB|}\sum_i\ell_i$.
                \ENDWHILE
                \STATE \textbf{return} $\bomega$.
            \end{algorithmic}
        \end{algorithm}
    \end{minipage}
    \hfill
    \begin{minipage}{0.49\linewidth}
        \begin{algorithm}[H]
            \caption{Training type II bridge networks}
            \label{alg:bridge_training_type2}
            \begin{algorithmic}
                \REQUIRE{Training dataset $\calD$, a pair of parameters $(\btheta_1, \btheta_2)$ and corresponding Bezier parameter $\bthetabe_{1,2}$,
                 a bridge network $\typetwo_{1,2}$ (type II) with parameters $\bomega$, learning rate $\eta$, a mixup coefficient $\alpha$.}
                \vspace{2mm}
                
                \STATE Initialize $\bomega$.
                \WHILE{not converged}
                    \STATE Sample a mini-batch $\calB \sim \calD$.
                    \FOR{$i=1,\dots,|\calB|$}
                        \STATE Take the input $\bsx_i$ from $\calB$.
                        \STATE $\tilde{\bsx_i} \gets \text{mixup}(\bsx_i, \alpha)$
                        \STATE $\bsz_1 \gets \fext_{\bphi_1}(\tilde{\bsx_i})$, $\bsv_1 \gets \fcls_{\bpsi_1}(\bsz_1)$.
                        \STATE $\bsz_2 \gets \fext_{\bphi_2}(\tilde{\bsx_i})$, $\bsv_2 \gets \fcls_{\bpsi_2}(\bsz_2)$.
                        \STATE $\bsv_{1,2}(r) \gets f_{\bthetabe_{1,2}(0.5)}(\tilde{\bsx_i})$.
                        \STATE $\tilde\bsv_{1,2}(r) = H_{1,2}^{(0.5)}(\bsz_1,\bsz_2;\bomega)$.
                        \STATE $\ell_i \gets D_\text{KL}(\bsv_{1,2}(0.5) || \tilde\bsv_{1,2}(0.5))$
                    \ENDFOR
                    \STATE $\bomega\gets \bomega - \eta \nabla_\bomega \frac{1}{|\calB|}\sum_i\ell_i$.
                \ENDWHILE
                \STATE \textbf{return} $\bomega$.
            \end{algorithmic}
        \end{algorithm}
    \end{minipage}
}

%%%%%%%%%%%%%%%%%%%%%%%%%%%%%%%%%%%%%%%%%%%%%%%%%%

\subsection{Filter Response Normalization}
Throughout experiments using convolutional neural networks, we use the Filter Response Normalization~\citep[FRN;][]{singh2020filter} instead of the Batch Normalization~\citep[BN;][]{ioffe2015batch} to avoid recomputation of BN statistics along the subspaces.
Besides, FRN is fully made up of learned parameters and it does not utilize dependencies between training examples, thus, it gives us a more clear interpretation of the parameter space~\citep{wenzel2020good,izmailov2021bayesian}.
We also perform experiments with Batch Normalization and present the results in \cref{sec:full_results_bn}.

%%%%%%%%%%%%%%%%%%%%%%%%%%%%%%%%%%%%%%%%%%%%%%%%%%

\subsection{Aligning the Bezier curves}
\label{sec:alignment}

In very difficult datasets such as ImageNet, it is not easy to reduce the training loss sufficiently small, and there are considerable discrepancies between trained models.
This makes the low-loss subspace between modes sufficiently complex, and the correlation between model outputs in the low-loss subspace becomes low.
A bridge network is built on the assumption that the outputs on the mode and the output on the low-loss subspace it approximates will be sufficiently correlated, so it will not perform well in this situation.
Therefore, in a situation where the base model does not have a sufficiently low training loss, as introduced in \citet{tatro2020optimizing}, we align the modes through permutation before finding a low-loss subspace between them.
Through this `neuron alignment', the correlation between the model output on the bezier curve and the model output in the mode rises significantly, and the bridge network works better.
Conversely, in relatively easy datasets like CIFAR-10, alignment reduces the diversity between low-loss subspaces and modes too much to help the ensemble.
Thus, we align models in mode before finding low-loss subspace on ImageNet dataset, and do not align on CIFAR-10/100, and Tiny ImageNet datasets.

%%%%%%%%%%%%%%%%%%%%%%%%%%%%%%%%%%%%%%%%%%%%%%%%%%

\subsection{Datasets and models}
\label{sec:train_setting}

\paragraph{Dataset}
We use CIFAR-10/100~\citep{Krizhevsky09learningmultiple}, Tiny ImageNet~\citep{tinyimagenet} and ImageNet~\citep{ILSVRC15} datasets.
We apply the data augmentation consisting of random cropping of 32 pixels with padding of 4 pixels and random horizontal flipping.
We subtract per-channel means from input images and divide them by per-channel standard deviations.

\paragraph{Network}
We employ similar ResNet block structures for the bridge networks in each dataset.
Each bridge network consists of three backbone blocks and one classifier layer.
To adjust FLOPs, we modify the channel sizes of the bridge network.
To utilize the features of the base models, we extract features $z$ from the third-to-last block of the base models.

\vspace{-1mm}
\begin{itemize}[leftmargin=*,itemsep=0mm]
    \item For CIFAR-10 dataset, we use ResNet-32$\times$2 as a base network which consists of 15 basic blocks (5, 5, 5) and 32 layers with widen factor of 2 and in-planes of 16.
    \item For CIFAR-100 dataset, we use ResNet-32$\times$4 as a base network which is almost the same network as CIFAR-10 with widen factor of 4.
    \item For Tiny ImageNet dataset, we use ResNet-34 as a base network which consists of 16 basic blocks (3, 4, 6, 3) and 34 layers with in-planes of 64.
    \item For ImageNet dataset, we use ResNet-50 as a base network which consists of 16 bottleneck blocks (3, 4, 6, 3) and 50 layers with in-planes of 64.
\end{itemize}

\paragraph{Optimization}
We train base ResNet networks for 200 epochs with learning rate 0.1.
We use the SGD optimizer with momentum 0.9 and adjust learning rate with simple cosine scheduler.
We give weight decay 0.001 for CIFAR-10 dataset, 0.0005 for CIFAR-100 and Tiny ImageNet dataset, and 0.0001 for ImageNet dataset.

\paragraph{Regularization}

We apply the mixup augmentation to train bridge models.
% We introduced two additional hyperparameters for training bridge models; 1) the regularization scale $\lambda$ and 2) the mixup coefficient $\alpha$.
Since the training error of the base network is near zero for the family of residual networks on CIFAR-10/100 and Tiny ImageNet, given a training input without any modification, the base network and the target network (the one on the Bezier curve) will produce almost identical outputs, so the bridge trained with them will just copy the outputs of the base network. To prevent this, we perturb the inputs via mixup.
% , and regularize the bridge to produce outputs different from the ones computed from the base models.
On the other hand, for the datasets such as ImageNet where the models fail to achieve near zero training errors, the base network and the target networks are already distinct enough, so we found that the bridge can be trained easily without such tricks (i.e., we used mixup coefficient $\alpha=0.0$).
We use $\alpha=0.4$ for CIFAR-10/100 and Tiny ImageNet datasets. We do not use mixup ($\alpha=0.0$) for ImageNet dataset.

%%%%%%%%%%%%%%%%%%%%%%%%%%%%%%%%%%%%%%%%%%%%%%%%%%

\subsection{Evaluation}
\label{sec:metrics}

\paragraph{Efficiency metrics}
\citet{dehghani2021efficiency} pointed out that there can be contradictions between commonly used metrics (e.g., FLOPs, the number of parameters, and speed) and suggested refraining from reporting results using just a single one. So, we present FLOPs and the number of parameters in the results.
% In terms of speed (e.g. latency, throughput), it is not easy to compare different types of models because this metric is highly dependent on the actual implementation. Considering the degree of parallelism, the overall computation time of \gls{de} may be slightly less than that of using a bridge network when much more computational power is used. However, the bridge network also can parallelize much of the computation and the overhead is expected to be minimal.

\paragraph{Uncertainty metrics}
Let $\bsp(\bsx) \in [0, 1]^K$ be a predicted probabilities for a given input $\bsx$, where $\bsp^{(k)}$ denotes the $k$th element of the probability vector, i.e., $\bsp^{(k)}$ is a predicted confidence on $k$th class.
We have the following common metrics on the dataset $\calD$ consists of inputs $\bsx$ and labels $y$:
\begin{itemize}[leftmargin=*]
    \item Accuracy (ACC):
    \vspace{-2mm}
    \[
    \operatorname{ACC}(\calD) = \bbE_{(\bsx,y)\in\calD} \left[ \left[
        y = \argmax_k \bsp^{(k)}(\bsx)
    \right] \right].
    \]
    \item Negative log-likelihood (NLL):
    \vspace{-2mm}
    \[
    \operatorname{NLL}(\calD) = \bbE_{(\bsx,y)\in\calD} \left[
        -\log{\bsp^{(y)}(\bsx)}
    \right].
    \]
    \item Brier score (BS):
    \vspace{-2mm}
    \[
    \operatorname{BS}(\calD) = \bbE_{(\bsx,y)\in\calD} \left[
        \Big\lVert \bsp(\bsx) - \bsy \Big\rVert_2^2
    \right],
    \]
    where $\bsy$ denotes one-hot encoded version of the label $y$, i.e., $\bsy^{(y)}=1$ and $\bsy^{(k)}=0$ for $k \neq y$.
    \item Expected calibration error (ECE):
    \vspace{-1mm}
    \[
    \operatorname{ECE}(\calD, N_{\text{bin}}) = \sum_{b=1}^{N_{\text{bin}}} \frac{
        n_b |\delta_b|
    }{
        n_1 + \cdots + n_{N_{\text{bin}}}
    },
    \]
    where $N_{\text{bin}}$ is the number of bins, $n_b$ is the number of examples in the $b$th bin, and $\delta_b$ is the calibration error of the $b$th bin.
    Specifically, the $b$th bin consists of predictions having the maximum confidence values in $[(b-1)/K, b/K)$, and the calibration error denotes the difference between accuracy and averaged confidences.
    We fix $N_{\text{bin}}=15$ in this paper.
\end{itemize}

We evaluate the \textit{calibrated} metrics that compute the aforementioned metrics with the temperature scaling~\citep{guo2017calibration}, as \citet{ashukha2020pitfalls} suggested.
Specifically, (1) we first find the optimal temperature which minimizes the NLL over the validation examples, and (2) compute uncertainty metrics including NLL, BS, and ECE using temperature scaled predicted probabilities under the optimal temperature.
Moreover, we evaluate the following Deep Ensemble Equivalent (DEE) score, which measure the relative performance for \gls{de} in terms of NLL,
\[
\operatorname{DEE}(\calD) = \min{\left\{ m \geq 0 \;|\; \operatorname{NLL}(\calD) \leq \operatorname{NLL}_{\text{DE-}m}(\calD) \right\}},
\]
where $\operatorname{NLL}_{\text{\gls{de}-}m}(\calD)$ denotes the NLL of \gls{de}-$m$ on the dataset $\calD$.
Here, we linearly interpolate $\operatorname{NLL}_{\text{\gls{de}-}m}(\calD)$ values for $m\in\bbR$ and make the DEE score continuous.

%%%%%%%%%%%%%%%%%%%%%%%%%%%%%%%%%%%%%%%%%%%%%%%%%%

\subsection{Computing Resources}
\label{sec:computing_resource}
We conduct Tiny ImageNet experiments on TPU v2-8 and TPU v3-8 supported by TPU Research Cloud\footnote{\href{https://sites.research.google/trc/about/}{https://sites.research.google/trc/about/}} and the others on NVIDIA GeForce RTX 3090.
We implemented the experimental codes using PyTorch~\citep{pytorch}.
\section{Additional Experiments}
\label{sec:additional_experiments}

%%%%%%%%%%%%%%%%%%%%%%%%%%%%%%%%%%%%%%%%%%%%%%%%%%

\subsection{Ablations on the choice of values for Bezier curve \textit{r}}
\label{sec:r_ablation}

\begin{table*}[h]
    \newcommand{\metricrule}{\cmidrule(lr){2-3} \cmidrule(lr){4-8}}
    \newcommand{\modelrule}{\cmidrule(lr){1-8}}
    \renewcommand{\arraystretch}{1.1}
    \centering
    \caption{
        $R^2$ score, KL divergence, and ensemble performances along Bezier curve $r \in (0, 1]$ on Tiny ImageNet dataset with Batch Normalization.
        Each type I bridge is trained to approximate the Bezier curve $r \in (0, 1]$ with 0.1 step size.
        $R^2$ score and KL divergence are calculated between the bridge output and the output of target Bezier curve $r$.
        Here, the type I bridge ($r=1.0$) actually tries to approximate the mode on the other end, and struggles to do so.
    }
    \vspace{-2mm}
    \resizebox{0.97\textwidth}{!}{
        \begin{tabular}{lllllllll}
            \textbf{Tiny ImageNet} \\
            \toprule
            Model                           & \umet{$R^2$}         & \dmet{KL}            & \umet{ACC}          & \dmet{NLL}           & \dmet{ECE}           & \dmet{BS}            & \umet{DEE}           \\
            \midrule
            % ResNet (\gls{de}-1)             & -                    & -                    & \tv[n]{62.66}{0.23} & \tv[n]{1.683}{0.009} & \tv[n]{0.050}{0.004} & \tv[n]{0.499}{0.002} & \tv[n]{1.000}{}      \\
            ResNet (\gls{de}-1)             & -                    & -                    & \tv[n]{63.90}{0.11} & \tv[n]{1.560}{0.006} & \tv[n]{0.035}{0.004} & \tv[n]{0.483}{0.002} & \tv[n]{1.000}{}      \\
            \metricrule
            \quad + Bridge\brgs \ ($r=0.1$) & \tv[n]{0.739}{0.003} & \tv[n]{0.600}{0.001} & \tv[n]{63.57}{0.24} & \tv[n]{1.533}{0.005} & \tv[n]{0.022}{0.002} & \tv[n]{0.485}{0.002} & \tv[n]{1.816}{0.042} \\
            \quad + Bridge\brgs \ ($r=0.2$) & \tv[n]{0.753}{0.002} & \tv[n]{0.539}{0.007} & \tv[n]{64.44}{0.12} & \tv[n]{1.508}{0.003} & \tv[n]{0.023}{0.004} & \tv[n]{0.472}{0.001} & \tv[n]{1.952}{0.061} \\
            \quad + Bridge\brgs \ ($r=0.3$) & \tv[n]{0.754}{0.005} & \tv[n]{0.511}{0.009} & \tv[n]{64.38}{0.20} & \tv[n]{1.507}{0.003} & \tv[n]{0.022}{0.000} & \tv[n]{0.472}{0.000} & \tv[n]{1.957}{0.055} \\
            \quad + Bridge\brgs \ ($r=0.4$) & \tv[n]{0.749}{0.007} & \tv[n]{0.506}{0.011} & \tv[n]{64.50}{0.15} & \tv[n]{1.494}{0.001} & \tv[n]{0.024}{0.001} & \tv[n]{0.470}{0.001} & \tv[n]{2.073}{0.085} \\
            \quad + Bridge\brgs \ ($r=0.5$) & \tv[b]{0.751}{0.008} & \tv[b]{0.498}{0.015} & \tv[n]{64.58}{0.17} & \tv[b]{1.478}{0.006} & \tv[n]{0.025}{0.002} & \tv[b]{0.469}{0.001} & \tv[b]{2.280}{0.086} \\
            \quad + Bridge\brgs \ ($r=0.6$) & \tv[n]{0.740}{0.013} & \tv[n]{0.530}{0.025} & \tv[b]{64.80}{0.17} & \tv[n]{1.483}{0.003} & \tv[n]{0.025}{0.002} & \tv[n]{0.470}{0.001} & \tv[n]{2.208}{0.053} \\
            \quad + Bridge\brgs \ ($r=0.7$) & \tv[n]{0.727}{0.012} & \tv[n]{0.576}{0.026} & \tv[n]{64.67}{0.33} & \tv[n]{1.494}{0.005} & \tv[n]{0.024}{0.001} & \tv[n]{0.470}{0.002} & \tv[n]{2.062}{0.116} \\
            \quad + Bridge\brgs \ ($r=0.8$) & \tv[n]{0.719}{0.011} & \tv[n]{0.626}{0.023} & \tv[n]{64.61}{0.15} & \tv[n]{1.492}{0.005} & \tv[n]{0.020}{0.002} & \tv[n]{0.470}{0.001} & \tv[n]{2.107}{0.116} \\
            \quad + Bridge\brgs \ ($r=0.9$) & \tv[n]{0.698}{0.008} & \tv[n]{0.721}{0.020} & \tv[n]{63.70}{0.17} & \tv[n]{1.526}{0.008} & \tv[n]{0.021}{0.004} & \tv[n]{0.484}{0.002} & \tv[n]{1.854}{0.022} \\
            \quad + Bridge\brgs \ ($r=1.0$) & \tv[n]{0.677}{0.008} & \tv[n]{0.810}{0.025} & \tv[n]{63.83}{0.10} & \tv[n]{1.529}{0.008} & \tv[b]{0.019}{0.003} & \tv[n]{0.484}{0.002} & \tv[n]{1.836}{0.052} \\
            % \modelrule
            % \gls{de}-2                      & -                    & -                    & \tv[n]{65.54}{0.25} & \tv[n]{1.499}{0.007} & \tv[n]{0.029}{0.003} & \tv[n]{0.461}{0.001} & \tv[n]{2.000}{}      \\
            % \gls{de}-2                      & -                    & -                    & \tv[n]{66.92}{0.22} & \tv[n]{1.401}{0.005} & \tv[n]{0.024}{0.003} & \tv[n]{0.444}{0.002} & \tv[n]{2.000}{}      \\
            \bottomrule
        \end{tabular}
    }
    \label{tab:table_r_ablation}
\end{table*}

We further examine the impact of selecting different values for a Bezier curve $r$ when constructing type I bridge networks.
In particular, $r=1.0$ would lead the type I bridge to directly predict the other mode from the given mode, which could have significant implications.
\cref{tab:table_r_ablation} summarizes the results on Tiny ImageNet dataset using ResNet with Batch Normalization.

Our proposed bridge networks are built on the hypothesis that there is a correspondence between two points on the low-loss Bezier curve in both weight and function space.
As demonstrated in \cref{sec:correspondence}, the type I bridge networks can accurately predict the function output of the target model ($r=0.5$) using the base model ($r=0.0$) due to the close relationship between the target and base models in terms of mode connectivity in weight space.

As we move further away from the base model (i.e., as $r$ increases), there could be further improvements.
While it is true for ideal bridge networks that perfectly predict the target model for any $r$ value in the range of $[0,1]$, we should note that the connection between the source and target models becomes weaker as we move away from the source.
The results clearly show that the type I bridge networks struggle in accurately predicting the target model with $r>0.6$, which is reflected in lower $R^2$ values.
As a result, the final ensemble performance suffers with lower ACC and higher NLL values.

%%%%%%%%%%%%%%%%%%%%%%%%%%%%%%%%%%%%%%%%%%%%%%%%%%

\subsection{Relationship between model size and regression result}
\label{sec:model_size_and_regression}

\begin{table*}[h]
    \renewcommand{\arraystretch}{1.1}
    \centering
    \caption{
        FLOPs, \#Params, $R^2$ scores, and \textit{ensemble} performance metrics of various type II bridge network sizes on CIFAR-100.
        We use ResNet-32$\times$4 as a base model and $3$ blocks of \gls{cnn} with a residual connection as bridge networks.
        The number after \gls{cnn} indicates the number of channels.
        $R^2$ scores are measured with respect to the target Bezier $r=0.5$.
    }
    \resizebox{0.9\textwidth}{!}{
        \begin{tabular}{lrrlllll}
            \toprule
            Bridge             & \dmet{FLOPs} & \dmet{\#Params} & \umet{$R^2$}         & \umet{ACC}          & \dmet{NLL}           & \dmet{ECE}           & \dmet{BS}            \\
            \midrule
            \gls{cnn} \; 32 ch & \relv{0.012} & \relv{0.009}    & \tv[n]{0.709}{0.004} & \tv[n]{75.62}{0.17} & \tv[n]{0.914}{0.005} & \tv[b]{0.013}{0.001} & \tv[n]{0.342}{0.002} \\
            \gls{cnn} \; 64 ch & \relv{0.029} & \relv{0.022}    & \tv[n]{0.758}{0.004} & \tv[n]{75.78}{0.30} & \tv[n]{0.901}{0.004} & \tv[n]{0.016}{0.002} & \tv[n]{0.338}{0.001} \\
            \gls{cnn}   128 ch & \relv{0.079} & \relv{0.060}    & \tv[n]{0.793}{0.003} & \tv[n]{75.98}{0.20} & \tv[n]{0.894}{0.003} & \tv[n]{0.021}{0.002} & \tv[n]{0.335}{0.001} \\
            \gls{cnn}   256 ch & \relv{0.252} & \relv{0.192}    & \tv[b]{0.805}{0.006} & \tv[b]{76.21}{0.11} & \tv[b]{0.863}{0.003} & \tv[n]{0.021}{0.002} & \tv[b]{0.324}{0.001} \\
            \bottomrule
        \end{tabular}
    }
    \label{tab:model_size}
\end{table*}

We measure the relationship between the size of the bridge networks and the goodness of predictions measured by $R^2$ scores.
\cref{tab:model_size} shows that we can achieve decent $R^2$ scores with a small number of parameters, and the performance improves as we increase the flexibility of our bridge network.

%%%%%%%%%%%%%%%%%%%%%%%%%%%%%%%%%%%%%%%%%%%%%%%%%%

\subsection{Using type I and type II bridges together}
\label{sec:type_mix}

\begin{table*}[ht]
    \newcommand{\metricrule}{\cmidrule(lr){2-3} \cmidrule(lr){4-8}}
    \newcommand{\modelrule}{\cmidrule(lr){1-8}}
    \renewcommand{\arraystretch}{1.1}
    \centering
    \caption{
        Performance improvement of the ensemble by adding both type I and type II bridges as members to existing DE ensembles on Tiny ImageNet datasets.
        FLOPs, \#Params, and DEE metrics are measured with respect to corresponding DEs.
        (I) denotes the bridge is type I, and (II) denotes the bridge is type II.
    }
    \vspace{-2mm}
    \parbox{0.68\textwidth}{%
        \resizebox{\linewidth}{!}{
            \begin{tabular}{lrrlllll}
                \textbf{Tiny ImageNet} \\
                \toprule
                Model                           & \dmet{FLOPs}  & \dmet{\#Params} & \umet{ACC}          & \dmet{NLL}           & \dmet{ECE}           & \dmet{BS}            & \umet{DEE}           \\
                \midrule
                \gls{de}-2                      & \relv{2.000} & \relv{2.000}     & \tv[n]{65.54}{0.25} & \tv[n]{1.499}{0.007} & \tv[n]{0.029}{0.003} & \tv[n]{0.461}{0.001} & \tv[n]{2.000}{}       \\
                \metricrule
                \quad + 1 Bridge\brgs (II)      & \relv{2.058} & \relv{2.067}     & \tv[n]{66.66}{0.37} & \tv[n]{1.394}{0.007} & \tv[n]{0.021}{0.002} & \tv[n]{0.445}{0.001} & \tv[n]{3.708}{0.187}  \\
                \quad \quad + 2 Bridge\brgs (I) & \relv{2.157} & \relv{2.181}     & \tv[b]{67.24}{0.19} & \tv[b]{1.341}{0.002} & \tv[b]{0.015}{0.002} & \tv[b]{0.437}{0.001} & \tv[b]{5.673}{0.219}  \\
                \modelrule
                \gls{de}-3                      & \relv{3.000} & \relv{3.000}     & \tv[n]{66.65}{0.18} & \tv[n]{1.425}{0.005} & \tv[n]{0.024}{0.002} & \tv[n]{0.444}{0.001} & \tv[n]{3.000}{}       \\
                \metricrule
                \quad + 3 Bridge\brgs (II)      & \relv{3.175} & \relv{3.202}     & \tv[n]{68.06}{0.26} & \tv[n]{1.312}{0.006} & \tv[n]{0.016}{0.001} & \tv[n]{0.428}{0.001} & \tv[n]{8.092}{0.610}  \\
                \quad \quad + 3 Bridge\brgs (I) & \relv{3.324} & \relv{3.373}     & \tv[b]{68.27}{0.17} & \tv[b]{1.292}{0.006} & \tv[b]{0.014}{0.002} & \tv[b]{0.427}{0.001} & \tv[b]{9.846}{0.718}  \\
                \modelrule
                \gls{de}-4                      & \relv{4.000} & \relv{4.000}     & \tv[n]{67.50}{0.11} & \tv[n]{1.381}{0.004} & \tv[n]{0.018}{0.001} & \tv[n]{0.435}{0.001} & \tv[n]{4.000}{}       \\
                \bottomrule
            \end{tabular}
        }
    }
    \label{tab:type_mix}
    \parbox{0.31\textwidth}{%
        \vspace{2mm}
        \resizebox{\linewidth}{!}{%
            \includegraphics{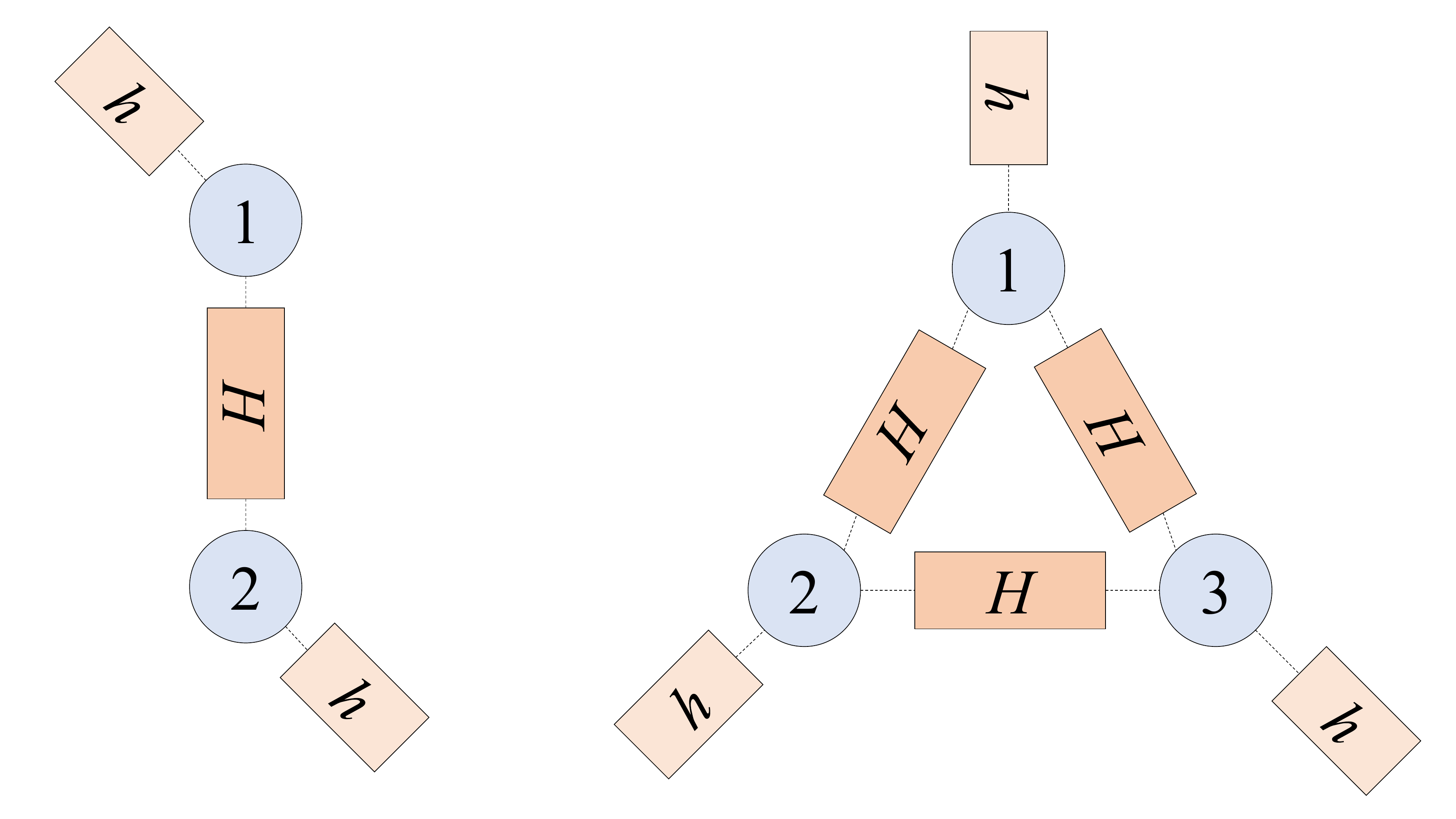}
        }
    }
    \vspace{-1mm}
    \captionof{figure}{
        `\gls{de}-2 + 2 Bridge\brgs (I) + 1 Bridge\brgs (II)' ensemble and `\gls{de}-3 + 3 Bridge\brgs (I) + 3 Bridge\brgs (II)' ensemble.
    }
    \label{fig:type_mix}
    \vspace{-2mm}
\end{table*}

Depending on the amount of computing resources available, we can use different combinations of bridge networks.
One can freely add more type I bridge networks to the ensemble of the base model and type II bridge networks.
As an example, let us assume that the computing resources can handle \gls{de}-2, but not \gls{de}-3.
Then we can use two base models to construct \gls{de}-2, and one type II bridge network connecting them.
After that, we can add more type I bridges to each base model to increase the performance of the ensemble.
There is no restriction to use only one type I bridge network on one base model, so more type I bridges can be used.
\cref{tab:type_mix} shows that adding two type I bridge networks further improves the performance of the `\gls{de}-2 + 1 Bridge\brgs (II)'.
And this can be similarly applied to the case of \gls{de}-3.

%%%%%%%%%%%%%%%%%%%%%%%%%%%%%%%%%%%%%%%%%%%%%%%%%%

\subsection{Selecting the optimal mode}
\label{sec:optimal_mode}

\begin{table*}[ht]
    \newcommand{\modelrule}{\cmidrule(lr){1-1} \cmidrule(lr){2-3} \cmidrule(lr){4-7} \cmidrule(lr){8-11}}
    \renewcommand{\arraystretch}{1.1}
    \centering
    \caption{
        Among the three modes with considerable and comparable performance differences, we choose one mode as the base endpoint and train type I bridges for the other two modes on Tiny ImageNet dataset.
        Selecting the mode with the best performance showed better ensemble performance than other cases.
    }
    \vspace{-2mm}
    \resizebox{0.85\textwidth}{!}{
        \begin{tabular}{lrrllllllll}
            \multicolumn{3}{l}{\textbf{Tiny ImageNet}} \\
            \toprule
                                            &              &                 & \multicolumn{4}{c}{Considerable differences}                          & \multicolumn{4}{c}{Comparable differences}                           \\
                                                                             \cmidrule(lr){4-7}                                                      \cmidrule(lr){8-11}
            Model                           & \dmet{FLOPs} & \dmet{\#Params} & \umet{ACC}      & \dmet{NLL}      & \dmet{ECE}      & \dmet{BS}       & \umet{ACC}      & \dmet{NLL}      & \dmet{ECE}      & \dmet{BS}       \\
            \midrule
            ResNet (High)                   & \relv{1.000} & \relv{1.000}    & \tv[n]{63.11}{} & \tv[n]{1.649}{} & \tv[n]{0.044}{} & \tv[n]{0.492}{} & \tv[n]{62.64}{} & \tv[n]{1.683}{} & \tv[n]{0.049}{} & \tv[n]{0.496}{} \\
            \quad + 2 Bridge\brgs           & \relv{1.099} & \relv{1.114}    & \tv[b]{65.90}{} & \tv[b]{1.415}{} & \tv[b]{0.014}{} & \tv[b]{0.455}{} & \tv[b]{65.48}{} & \tv[b]{1.421}{} & \tv[b]{0.023}{} & \tv[b]{0.457}{} \\
            \modelrule
            ResNet (Center)                 & \relv{1.000} & \relv{1.000}    & \tv[n]{62.64}{} & \tv[n]{1.683}{} & \tv[n]{0.049}{} & \tv[n]{0.496}{} & \tv[n]{62.61}{} & \tv[n]{1.687}{} & \tv[n]{0.053}{} & \tv[n]{0.501}{} \\
            \quad + 2 Bridge\brgs           & \relv{1.099} & \relv{1.114}    & \tv[n]{65.45}{} & \tv[n]{1.421}{} & \tv[n]{0.020}{} & \tv[n]{0.458}{} & \tv[n]{65.43}{} & \tv[n]{1.427}{} & \tv[n]{0.020}{} & \tv[n]{0.459}{} \\
            \modelrule
            ResNet (Low)                    & \relv{1.000} & \relv{1.000}    & \tv[n]{62.38}{} & \tv[n]{1.693}{} & \tv[n]{0.053}{} & \tv[n]{0.501}{} & \tv[n]{62.38}{} & \tv[n]{1.693}{} & \tv[n]{0.053}{} & \tv[n]{0.501}{} \\
            \quad + 2 Bridge\brgs           & \relv{1.099} & \relv{1.114}    & \tv[n]{65.29}{} & \tv[n]{1.419}{} & \tv[n]{0.021}{} & \tv[n]{0.458}{} & \tv[n]{65.09}{} & \tv[n]{1.417}{} & \tv[n]{0.020}{} & \tv[n]{0.457}{} \\
            \bottomrule
        \end{tabular}
    }
    \label{tab:optimal_mode}
    \vspace{-2mm}
\end{table*}

Given the practicality of the proposed type I bridge networks, we investigate how to choose the base model among multiple modes.
To this end, we consider three modes and evaluate the performance of ``\gls{de}-1 + 2 type I Bridge\brgs'' using three different base models.
\cref{tab:optimal_mode} briefly demonstrates this scenario.
As a result, selecting a good-performing base model generally leads to good ensemble performance.
However, in cases where the performance difference between modes is not significant, the difference in ensemble performance was also not significantly different within the margin of error. 

%%%%%%%%%%%%%%%%%%%%%%%%%%%%%%%%%%%%%%%%%%%%%%%%%%

\subsection{Using bridge networks with SWA trained base model}
\label{sec:swa_bridge}

\vspace{-1mm}

\begin{table*}[h]
    \newcommand{\metricrule}{\cmidrule(lr){2-3} \cmidrule(lr){4-7}}
    \newcommand{\modelrule}{\cmidrule(lr){1-7}}
    \newcommand{\methodrule}{\cmidrule(lr){1-1} \cmidrule(lr){2-3} \cmidrule(lr){4-7}}
    \centering
    \caption{
        Comparison of performance improvement of the efficient methods on Tiny ImageNet dataset.
        FLOPs, \#Params, and DEE metrics are measured with respect to the single ResNet-34.
    }
    \vspace{-5mm}
    \parbox{0.64\textwidth}{%
        \resizebox{\linewidth}{!}{%
            \begin{tabular}{lrrlllll}
                \textbf{Tiny ImageNet} \\
                \toprule
                Model                    & \dmet{FLOPs} & \dmet{\#Params} & \umet{ACC}          & \dmet{NLL}           & \dmet{ECE}           & \umet{DEE}           \\
                \midrule
                ResNet (\small{SWA})     & \relv{1.000} & \relv{1.000}    & \tv[n]{64.03}{0.21} & \tv[n]{1.519}{0.010} & \tv[n]{0.030}{0.002} & \tv[n]{1.888}{0.074} \\
                \metricrule
                \quad + 1 Bridge\brgs    & \relv{1.050} & \relv{1.057}    & \tv[n]{65.26}{0.08} & \tv[n]{1.435}{0.004} & \tv[n]{0.031}{0.001} & \tv[n]{2.865}{0.087} \\
                \quad + 2 Bridge\brgs    & \relv{1.099} & \relv{1.114}    & \tv[n]{65.77}{0.10} & \tv[n]{1.403}{0.003} & \tv[n]{0.028}{0.002} & \tv[n]{3.511}{0.162} \\
                \quad + 3 Bridge\brgs    & \relv{1.149} & \relv{1.171}    & \tv[n]{65.96}{0.09} & \tv[n]{1.387}{0.001} & \tv[b]{0.026}{0.002} & \tv[n]{3.873}{0.160} \\
                \metricrule
                \quad + 1 Bridge\brgm    & \relv{1.180} & \relv{1.206}    & \tv[n]{65.46}{0.05} & \tv[n]{1.422}{0.004} & \tv[n]{0.030}{0.001} & \tv[n]{3.093}{0.173} \\
                \quad + 2 Bridge\brgm    & \relv{1.359} & \relv{1.412}    & \tv[n]{66.06}{0.15} & \tv[n]{1.386}{0.004} & \tv[n]{0.025}{0.002} & \tv[n]{3.903}{0.190} \\
                \quad + 3 Bridge\brgm    & \relv{1.539} & \relv{1.618}    & \tv[n]{66.46}{0.14} & \tv[n]{1.369}{0.002} & \tv[n]{0.024}{0.002} & \tv[n]{4.431}{0.181} \\
                \quad + 4 Bridge\brgm    & \relv{1.719} & \relv{1.824}    & \tv[b]{66.71}{0.04} & \tv[b]{1.359}{0.001} & \tv[b]{0.022}{0.001} & \tv[b]{4.817}{0.111} \\
                \modelrule
                \gls{de}-2 (\small{SWA}) & \relv{2.000} & \relv{2.000}    & \tv[n]{66.28}{0.07} & \tv[n]{1.400}{0.003} & \tv[n]{0.020}{0.002} & \tv[n]{3.565}{0.156} \\
                \bottomrule
            \end{tabular}
        }
    }
    \label{tab:swa_bridge}
    \parbox{0.35\textwidth}{%
        \vspace{2mm}
        \resizebox{\linewidth}{!}{%
            \includegraphics{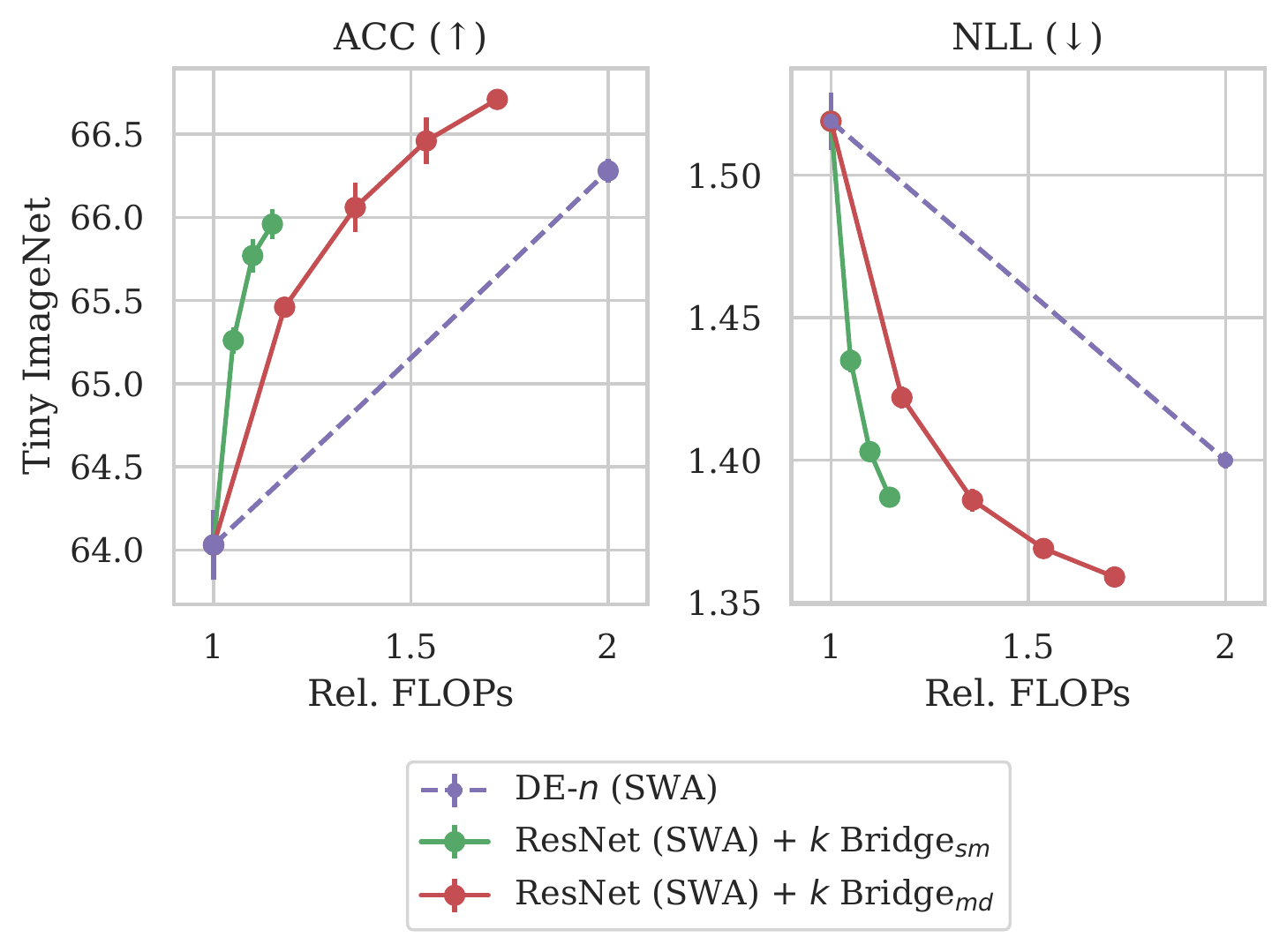}
        }
    }
    \vspace{-2mm}
    \captionof{figure}{
        The cost-performance plots of efficient methods compared to \gls{de} of ResNet-34 on Tiny ImageNet dataset.
        The x-axis denotes the relative FLOPs quantifying the inference cost of the model compared to a single ResNet-34, and the y-axis shows the corresponding predictive performance.
        On the basis of \gls{de} (black dashed line), the upper left position is preferable in ACC, and the lower left position is preferable in NLL.
    }
    \label{fig:swa_bridge}
    \vspace{-4mm}
\end{table*}

We trained base models with SWA and also train Bezier curves connecting them.
Then we train bridge networks to learn curves using features from SWA-trained base models.
\cref{tab:swa_bridge} and \cref{fig:swa_bridge} show that the bridge networks also give performance improvements when the base models are trained with SWA.

%%%%%%%%%%%%%%%%%%%%%%%%%%%%%%%%%%%%%%%%%%%%%%%%%%

\subsection{Additional examples}
\label{sec:additional_examples}

\begin{figure*}[t]
    \centering
    \includegraphics[width=0.98\textwidth]{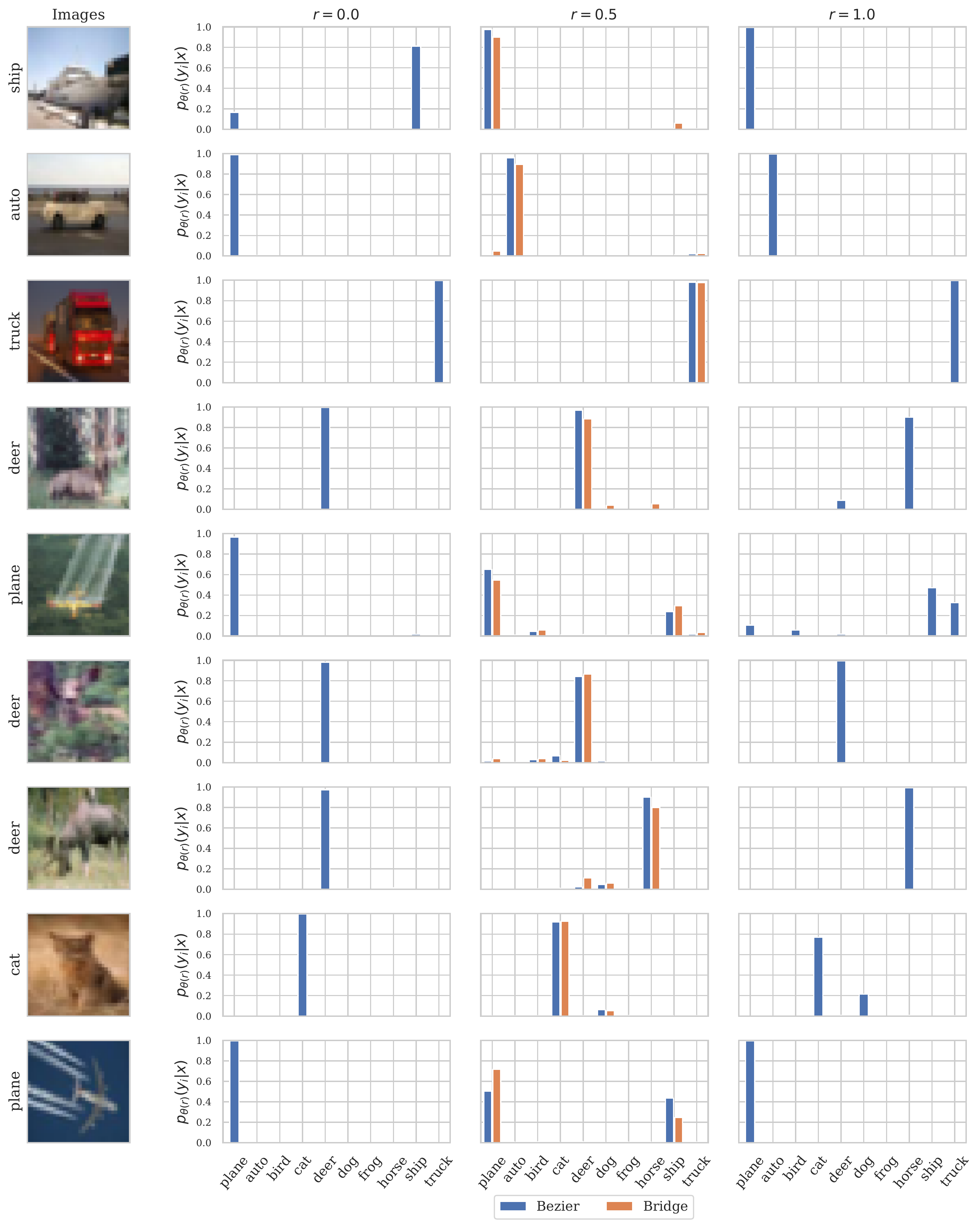}
    \caption{
    Bar plots in the third column depict whether the bridge network (\textbf{orange}) outputs the same class probability values as the base model with the Bezier parameters $\bthetabe_{1,2}(0.5)$ (\textbf{blue}), for a given test inputs displayed in the first column. We also depict the predicted logits from $\btheta_1$ and $\btheta_2$ in the second and fourth columns, respectively.
    }
    \label{fig:logit_regression_full}
\end{figure*}

In \cref{fig:logit_regression_full}, we visually inspect the logit regression of a type II bridge network.
Our bridge network very accurately predicts the logits of $r=0.5$ from Bezier curve when the two base models ($r=0$ and $r=1$) gives similar output logits (deer, ship, and frog).
When the base models are not confident on the samples (airplane, bird, cat, and horse), the network recovers the scale of logits approximately.
However, it fails to predict some challenging samples (truck and dog) when even the base models are very confused.

%%%%%%%%%%%%%%%%%%%%%%%%%%%%%%%%%%%%%%%%%%%%%%%%%%

\subsection{Full type I and type II bridge results}
\label{sec:full_results}

We report full experimental results for classification tasks; (1) Type I bridge network results in \cref{tab:type1_c10}, \cref{tab:type1_c100}, \cref{tab:type1_t200} and \cref{tab:type1_in1k}, (2) Type II bridge network results in \cref{tab:type2_c10}, \cref{tab:type2_c100}, \cref{tab:type2_t200} and \cref{tab:type2_in1k}.

%%%%%%%%%%%%%%%%%%%%%%%%%%%%%%%%%%%%%%%%%%%%%%%%%%

\subsection{Full type I and type II bridge results with Batch Normalization}
\label{sec:full_results_bn}

We report complete experimental results for classification tasks with Batch Normalization instead of Filter Response Normalization; (1) Type I bridge network results in \cref{tab:type1_c10_bn}, \cref{tab:type1_c100_bn} and \cref{tab:type1_t200_bn}, (2) Type II bridge network results in \cref{tab:type2_c10_bn}, \cref{tab:type2_c100_bn} and \cref{tab:type2_t200_bn}.
The results indicate that the choice of a normalization operation is an architectural detail, and our approach is also compatible with conventional Batch Normalization.

%%%%%%%%%%%%%%%%%%%%%%%%%%%%%%%%%%%%%%%%%%%%%%%%%%

\clearpage
\begin{table*}[t]
    \newcommand{\metricrule}{\cmidrule(lr){2-3} \cmidrule(lr){4-8}}
    \newcommand{\modelrule}{\cmidrule(lr){1-8}}
    \renewcommand{\arraystretch}{1.18}
    \centering
    \caption{
        Full result of performance improvement of the ensemble by adding type I bridges on CIFAR-10 dataset.
        We use the same settings as described in \cref{tab:type1}.
    }
    \vspace{-2mm}
    \resizebox{0.9\textwidth}{!}{
\begin{tabular}{lrrllllll}
            \textbf{CIFAR-10} \\
            \toprule
            Model                 & \dmet{FLOPs} & \dmet{\#Params} & \umet{ACC}          & \dmet{NLL}           & \dmet{ECE}           & \dmet{BS}            & \umet{DEE}           \\
            \midrule
            ResNet (\gls{de}-1)   & \relv{1.000} & \relv{1.000}    & \tv[n]{91.78}{0.10} & \tv[n]{0.287}{0.001} & \tv[n]{0.019}{0.001} & \tv[n]{0.126}{0.001} & \tv[n]{1.000}{}      \\
            \metricrule
            \quad + 1 Bridge\brgs & \relv{1.062} & \relv{1.048}    & \tv[n]{92.10}{0.04} & \tv[n]{0.252}{0.001} & \tv[n]{0.011}{0.001} & \tv[n]{0.118}{0.000} & \tv[n]{1.645}{0.041} \\
            \quad + 2 Bridge\brgs & \relv{1.125} & \relv{1.097}    & \tv[n]{92.15}{0.04} & \tv[n]{0.246}{0.001} & \tv[b]{0.009}{0.000} & \tv[n]{0.117}{0.000} & \tv[n]{1.759}{0.051} \\
            \quad + 3 Bridge\brgs & \relv{1.187} & \relv{1.145}    & \tv[n]{92.18}{0.04} & \tv[n]{0.244}{0.001} & \tv[n]{0.011}{0.000} & \tv[b]{0.116}{0.000} & \tv[n]{1.801}{0.055} \\
            \quad + 4 Bridge\brgs & \relv{1.249} & \relv{1.194}    & \tv[b]{92.22}{0.06} & \tv[b]{0.243}{0.000} & \tv[n]{0.010}{0.001} & \tv[b]{0.116}{0.000} & \tv[b]{1.819}{0.046} \\
            \metricrule
            \quad + 1 Bridge\brgm & \relv{1.205} & \relv{1.159}    & \tv[n]{92.09}{0.08} & \tv[n]{0.251}{0.001} & \tv[n]{0.011}{0.000} & \tv[n]{0.118}{0.000} & \tv[n]{1.674}{0.029} \\
\bottomrule
        \end{tabular}
    }
    \label{tab:type1_c10}
\end{table*}

\begin{table*}[t]
    \newcommand{\metricrule}{\cmidrule(lr){2-3} \cmidrule(lr){4-8}}
    \newcommand{\modelrule}{\cmidrule(lr){1-8}}
    \renewcommand{\arraystretch}{1.18}
    \centering
    \caption{
        Full result of performance improvement of the ensemble by adding type II bridges on CIFAR-10 dataset.
        We use the same settings as described in \cref{tab:type2}.
    }
    \vspace{-2mm}
    \resizebox{0.9\textwidth}{!}{
        \begin{tabular}{lrrlllll}
            \textbf{CIFAR-10} \\
            \toprule
            Model                 & \dmet{FLOPs}  & \dmet{\#Params} & \umet{ACC}          & \dmet{NLL}           & \dmet{ECE}           & \dmet{BS}            & \umet{DEE}           \\
            \midrule
            \gls{de}-2            & \relv{ 2.000} & \relv{ 2.000}   & \tv[n]{93.07}{0.11} & \tv[n]{0.233}{0.003} & \tv[n]{0.012}{0.001} & \tv[n]{0.107}{0.001} & \tv[n]{2.000}{}      \\
            \metricrule
            \quad + 1 Bridge\brgs & \relv{ 2.081} & \relv{ 2.063}   & \tv[b]{93.09}{0.09} & \tv[b]{0.219}{0.003} & \tv[b]{0.007}{0.001} & \tv[b]{0.103}{0.001} & \tv[b]{2.811}{0.143} \\
            \metricrule
            \quad + 1 Bridge\brgm & \relv{ 2.247} & \relv{ 2.192}   & \tv[b]{93.09}{0.15} & \tv[b]{0.219}{0.003} & \tv[b]{0.007}{0.001} & \tv[b]{0.103}{0.001} & \tv[b]{2.837}{0.160} \\
            \metricrule
            \quad + 1 Bezier      & \relv{ 3.000} & \relv{ 3.000}   & \tv[n]{93.03}{0.11} & \tv[n]{0.227}{0.004} & \tv[n]{0.008}{0.001} & \tv[n]{0.105}{0.001} & \tv[n]{2.377}{0.090} \\
            \modelrule
            \gls{de}-3            & \relv{ 3.000} & \relv{ 3.000}   & \tv[n]{93.37}{0.03} & \tv[n]{0.216}{0.004} & \tv[n]{0.009}{0.001} & \tv[n]{0.100}{0.001} & \tv[n]{3.000}{}      \\
            \metricrule
            \quad + 1 Bridge\brgs & \relv{ 3.081} & \relv{ 3.063}   & \tv[n]{93.45}{0.08} & \tv[n]{0.206}{0.002} & \tv[b]{0.006}{0.002} & \tv[n]{0.097}{0.001} & \tv[n]{3.896}{0.115} \\
\bottomrule
        \end{tabular}
    }
    \label{tab:type2_c10}
\end{table*}

\clearpage
\begin{table*}[t]
    \newcommand{\metricrule}{\cmidrule(lr){2-3} \cmidrule(lr){4-8}}
    \newcommand{\modelrule}{\cmidrule(lr){1-8}}
    \renewcommand{\arraystretch}{1.18}
    \centering
    \caption{
        Full result of performance improvement of the ensemble by adding type I bridges on CIFAR-100 dataset.
        We use the same settings as described in \cref{tab:type1}.
    }
    \vspace{-2mm}
    \resizebox{0.9\textwidth}{!}{
        \begin{tabular}{lrrlllll}
            \textbf{CIFAR-100} \\
            \toprule
            Model                 & \dmet{FLOPs} & \dmet{\#Params} & \umet{ACC}          & \dmet{NLL}           & \dmet{ECE}           & \dmet{BS}            & \umet{DEE}           \\
            \midrule
            ResNet (\gls{de}-1)   & \relv{1.000} & \relv{1.000}    & \tv[n]{74.06}{0.31} & \tv[n]{1.060}{0.005} & \tv[n]{0.047}{0.003} & \tv[n]{0.367}{0.002} & \tv[n]{1.000}{}      \\
            \metricrule
            \quad + 1 Bridge\brgs & \relv{1.063} & \relv{1.048}    & \tv[n]{75.03}{0.07} & \tv[n]{0.938}{0.001} & \tv[n]{0.029}{0.001} & \tv[n]{0.347}{0.000} & \tv[n]{1.928}{0.038} \\
            \quad + 2 Bridge\brgs & \relv{1.127} & \relv{1.097}    & \tv[n]{75.47}{0.11} & \tv[n]{0.911}{0.001} & \tv[n]{0.025}{0.000} & \tv[n]{0.342}{0.001} & \tv[n]{2.321}{0.076} \\
            \quad + 3 Bridge\brgs & \relv{1.190} & \relv{1.145}    & \tv[b]{75.58}{0.16} & \tv[n]{0.900}{0.001} & \tv[n]{0.025}{0.001} & \tv[n]{0.340}{0.000} & \tv[n]{2.545}{0.077} \\
            \quad + 4 Bridge\brgs & \relv{1.253} & \relv{1.194}    & \tv[b]{75.58}{0.07} & \tv[b]{0.893}{0.001} & \tv[b]{0.024}{0.001} & \tv[b]{0.339}{0.000} & \tv[b]{2.673}{0.070} \\
            \metricrule
            \quad + 1 Bridge\brgm & \relv{1.210} & \relv{1.160}    & \tv[n]{75.25}{0.09} & \tv[n]{0.929}{0.003} & \tv[n]{0.033}{0.001} & \tv[n]{0.344}{0.001} & \tv[n]{2.008}{0.042} \\
\bottomrule
        \end{tabular}
    }
    \label{tab:type1_c100}
\end{table*}

\begin{table*}[t]
    \newcommand{\metricrule}{\cmidrule(lr){2-3} \cmidrule(lr){4-8}}
    \newcommand{\modelrule}{\cmidrule(lr){1-8}}
    \renewcommand{\arraystretch}{1.18}
    \centering
    \caption{
        Full result of performance improvement of the ensemble by adding type II bridges on CIFAR-100 dataset.
        We use the same settings as described in \cref{tab:type2}.
    }
    \vspace{-2mm}
    \resizebox{0.9\textwidth}{!}{
        \begin{tabular}{lrrlllll}
            \textbf{CIFAR-100} \\
            \toprule
            Model                 & \dmet{FLOPs}  & \dmet{\#Params} & \umet{ACC}          & \dmet{NLL}           & \dmet{ECE}           & \dmet{BS}            & \umet{DEE}           \\
            \midrule
            \gls{de}-2            & \relv{ 2.000} & \relv{ 2.000}   & \tv[n]{76.21}{0.11} & \tv[n]{0.928}{0.004} & \tv[n]{0.027}{0.003} & \tv[n]{0.334}{0.001} & \tv[n]{2.000}{}      \\
            \metricrule
            \quad + 1 Bridge\brgs & \relv{ 2.083} & \relv{ 2.063}   & \tv[b]{76.72}{0.19} & \tv[b]{0.868}{0.001} & \tv[b]{0.025}{0.001} & \tv[b]{0.325}{0.001} & \tv[b]{3.256}{0.096} \\
            \metricrule
            \quad + 1 Bridge\brgm & \relv{ 2.252} & \relv{ 2.192}   & \tv[b]{76.75}{0.12} & \tv[b]{0.863}{0.003} & \tv[b]{0.024}{0.002} & \tv[b]{0.324}{0.001} & \tv[b]{3.435}{0.065} \\
            \metricrule
            \quad + 1 Bezier      & \relv{ 3.000} & \relv{ 3.000}   & \tv[n]{76.87}{0.10} & \tv[n]{0.879}{0.005} & \tv[n]{0.021}{0.002} & \tv[n]{0.324}{0.001} & \tv[n]{2.954}{0.111} \\
            \modelrule
            \gls{de}-3            & \relv{ 3.000} & \relv{ 3.000}   & \tv[n]{77.23}{0.18} & \tv[n]{0.876}{0.003} & \tv[n]{0.024}{0.001} & \tv[n]{0.321}{0.001} & \tv[n]{3.000}{}      \\
            \metricrule
            \quad + 1 Bridge\brgs & \relv{ 3.083} & \relv{ 3.063}   & \tv[n]{77.43}{0.09} & \tv[n]{0.838}{0.001} & \tv[n]{0.021}{0.002} & \tv[n]{0.316}{0.001} & \tv[n]{4.429}{0.112} \\
\bottomrule
        \end{tabular}
    }
    \label{tab:type2_c100}
\end{table*}

\clearpage
\begin{table*}[t]
    \newcommand{\metricrule}{\cmidrule(lr){2-3} \cmidrule(lr){4-8}}
    \newcommand{\modelrule}{\cmidrule(lr){1-8}}
    \renewcommand{\arraystretch}{1.18}
    \centering
    \caption{
        Full result of performance improvement of the ensemble by adding type I bridges on Tiny ImageNet dataset.
        We use the same settings as described in \cref{tab:type1}.
    }
    \vspace{-2mm}
    \resizebox{0.9\textwidth}{!}{
        \begin{tabular}{lrrlllll}
            \textbf{Tiny ImageNet} \\
            \toprule
            Model                 & \dmet{FLOPs} & \dmet{\#Params} & \umet{ACC}          & \dmet{NLL}           & \dmet{ECE}           & \dmet{BS}            & \umet{DEE}           \\
            \midrule
            ResNet (\gls{de}-1)   & \relv{1.000} & \relv{1.000}    & \tv[n]{62.66}{0.23} & \tv[n]{1.683}{0.009} & \tv[n]{0.050}{0.004} & \tv[n]{0.499}{0.002} & \tv[n]{1.000}{}      \\
            \metricrule
            \quad + 1 Bridge\brgs & \relv{1.050} & \relv{1.057}    & \tv[n]{64.58}{0.17} & \tv[n]{1.478}{0.006} & \tv[n]{0.025}{0.002} & \tv[n]{0.469}{0.001} & \tv[n]{2.280}{0.086} \\
            \quad + 2 Bridge\brgs & \relv{1.099} & \relv{1.114}    & \tv[n]{65.37}{0.13} & \tv[n]{1.421}{0.004} & \tv[n]{0.018}{0.002} & \tv[n]{0.459}{0.000} & \tv[n]{3.087}{0.118} \\
            \quad + 3 Bridge\brgs & \relv{1.149} & \relv{1.171}    & \tv[n]{65.82}{0.10} & \tv[n]{1.395}{0.003} & \tv[n]{0.015}{0.001} & \tv[n]{0.454}{0.001} & \tv[n]{3.680}{0.133} \\
            \quad + 4 Bridge\brgs & \relv{1.198} & \relv{1.228}    & \tv[b]{65.93}{0.10} & \tv[b]{1.380}{0.001} & \tv[b]{0.014}{0.002} & \tv[b]{0.452}{0.000} & \tv[b]{4.061}{0.157} \\
            \metricrule
            \quad + 1 Bridge\brgm & \relv{1.180} & \relv{1.206}    & \tv[n]{65.13}{0.12} & \tv[n]{1.446}{0.002} & \tv[n]{0.034}{0.002} & \tv[n]{0.461}{0.001} & \tv[n]{2.709}{0.049} \\
\bottomrule
        \end{tabular}
    }
    \label{tab:type1_t200}
\end{table*}

\begin{table*}[t]
    \newcommand{\metricrule}{\cmidrule(lr){2-3} \cmidrule(lr){4-8}}
    \newcommand{\modelrule}{\cmidrule(lr){1-8}}
    \renewcommand{\arraystretch}{1.18}
    \newcommand{\hs}{\hspace{1.8mm}}
    \centering
    \caption{
        Full result of performance improvement of the ensemble by adding type II bridges on Tiny ImageNet dataset.
        We use the same settings as described in \cref{tab:type2}.
    }
    \vspace{-2mm}
    \resizebox{0.9\textwidth}{!}{
        \begin{tabular}{lrrlllll}
            \textbf{Tiny ImageNet} \\
            \toprule
            Model                 & \dmet{FLOPs}  & \dmet{\#Params} & \umet{ACC}          & \dmet{NLL}           & \dmet{ECE}           & \dmet{BS}            & \umet{DEE}           \\
            \midrule
            \gls{de}-2            & \relv{ 2.000} & \relv{ 2.000}   & \tv[n]{65.54}{0.25} & \tv[n]{1.499}{0.007} & \tv[n]{0.029}{0.003} & \tv[n]{0.461}{0.001} & \tv[n]{\hs2.000}{}      \\
            \metricrule
            \quad + 1 Bridge\brgs & \relv{ 2.058} & \relv{ 2.067}   & \tv[b]{66.66}{0.37} & \tv[b]{1.394}{0.007} & \tv[b]{0.021}{0.002} & \tv[b]{0.445}{0.001} & \tv[b]{\hs3.708}{0.187} \\
            \metricrule
            \quad + 1 Bridge\brgm & \relv{ 2.198} & \relv{ 2.226}   & \tv[b]{66.72}{0.15} & \tv[b]{1.386}{0.005} & \tv[b]{0.025}{0.001} & \tv[b]{0.443}{0.001} & \tv[b]{\hs3.901}{0.109} \\
            \metricrule
            \quad + 1 Bezier      & \relv{ 3.000} & \relv{ 3.000}   & \tv[n]{66.17}{0.12} & \tv[n]{1.436}{0.004} & \tv[n]{0.024}{0.001} & \tv[n]{0.449}{0.001} & \tv[n]{\hs2.851}{0.059} \\
            \modelrule
            \gls{de}-3            & \relv{ 3.000} & \relv{ 3.000}   & \tv[n]{66.65}{0.18} & \tv[n]{1.425}{0.005} & \tv[n]{0.024}{0.002} & \tv[n]{0.444}{0.001} & \tv[n]{\hs3.000}{}      \\
            \metricrule
            \quad + 1 Bridge\brgs & \relv{ 3.058} & \relv{ 3.067}   & \tv[n]{67.38}{0.18} & \tv[n]{1.358}{0.005} & \tv[b]{0.016}{0.001} & \tv[n]{0.435}{0.001} & \tv[n]{\hs4.868}{0.189} \\
\bottomrule
        \end{tabular}
    }
    \label{tab:type2_t200}
\end{table*}

\clearpage
\begin{table*}[t]
    \newcommand{\metricrule}{\cmidrule(lr){2-3} \cmidrule(lr){4-8}}
    \newcommand{\modelrule}{\cmidrule(lr){1-8}}
    \renewcommand{\arraystretch}{1.18}
    \centering
    \caption{
        Full result of performance improvement of the ensemble by adding type I bridges on ImageNet dataset.
        We use the same settings as described in \cref{tab:type1}.
    }
    \vspace{-2mm}
    \resizebox{0.9\textwidth}{!}{
        \begin{tabular}{lrrlllll}
            \textbf{ImageNet} \\
            \toprule
            Model                 & \dmet{FLOPs} & \dmet{\#Params} & \umet{ACC}          & \dmet{NLL}           & \dmet{ECE}           & \dmet{BS}            & \umet{DEE}           \\
            \midrule
            ResNet (\gls{de}-1)   & \relv{1.000} & \relv{1.000}    & \tv[n]{75.85}{0.06} & \tv[n]{0.936}{0.003} & \tv[n]{0.019}{0.001} & \tv[n]{0.333}{0.001} & \tv[n]{1.000}{}      \\
            \metricrule
            \quad + 1 Bridge\brgs & \relv{1.061} & \relv{1.071}    & \tv[n]{76.46}{0.06} & \tv[n]{0.914}{0.000} & \tv[n]{0.012}{0.001} & \tv[n]{0.326}{0.000} & \tv[n]{1.418}{0.034} \\
            \quad + 2 Bridge\brgs & \relv{1.123} & \relv{1.141}    & \tv[n]{76.60}{0.06} & \tv[n]{0.907}{0.000} & \tv[n]{0.012}{0.001} & \tv[n]{0.324}{0.000} & \tv[n]{1.537}{0.026} \\
            \quad + 3 Bridge\brgs & \relv{1.184} & \relv{1.212}    & \tv[n]{76.69}{0.04} & \tv[n]{0.905}{0.000} & \tv[b]{0.011}{0.001} & \tv[n]{0.323}{0.000} & \tv[n]{1.584}{0.021} \\
            \quad + 4 Bridge\brgs & \relv{1.245} & \relv{1.283}    & \tv[b]{76.71}{0.03} & \tv[b]{0.904}{0.000} & \tv[b]{0.011}{0.001} & \tv[b]{0.322}{0.000} & \tv[b]{1.610}{0.021} \\
            \metricrule
            \quad + 1 Bridge\brgm & \relv{1.194} & \relv{1.222}    & \tv[n]{77.03}{0.07} & \tv[n]{0.889}{0.001} & \tv[b]{0.013}{0.000} & \tv[n]{0.319}{0.000} & \tv[n]{1.881}{0.022} \\
\bottomrule
        \end{tabular}
    }
    \label{tab:type1_in1k}
\end{table*}

\begin{table*}[t]
    \newcommand{\metricrule}{\cmidrule(lr){2-3} \cmidrule(lr){4-8}}
    \newcommand{\modelrule}{\cmidrule(lr){1-8}}
    \renewcommand{\arraystretch}{1.18}
    \newcommand{\hs}{\hspace{1.8mm}}
    \centering
    \caption{
        Full result of performance improvement of the ensemble by adding type II bridges on ImageNet dataset.
        We use the same settings as described in \cref{tab:type2}.
    }
    \vspace{-2mm}
    \resizebox{0.9\textwidth}{!}{
        \begin{tabular}{lrrlllll}
            \textbf{ImageNet} \\
            \toprule
            Model                 & \dmet{FLOPs}  & \dmet{\#Params} & \umet{ACC}          & \dmet{NLL}           & \dmet{ECE}           & \dmet{BS}            & \umet{DEE}           \\
            \midrule
            \gls{de}-2            & \relv{ 2.000} & \relv{ 2.000}   & \tv[n]{77.12}{0.04} & \tv[n]{0.883}{0.001} & \tv[n]{0.012}{0.001} & \tv[n]{0.318}{0.000} & \tv[n]{\hs2.000}{}      \\
            \metricrule
            \quad + 1 Bridge\brgs & \relv{ 2.086} & \relv{ 2.088}   & \tv[b]{77.44}{0.03} & \tv[b]{0.871}{0.001} & \tv[b]{0.012}{0.000} & \tv[b]{0.314}{0.000} & \tv[b]{\hs2.567}{0.017} \\
            \metricrule
            \quad + 1 Bridge\brgm & \relv{ 2.243} & \relv{ 2.256}   & \tv[b]{77.70}{0.05} & \tv[b]{0.857}{0.001} & \tv[b]{0.012}{0.001} & \tv[b]{0.310}{0.000} & \tv[b]{\hs3.416}{0.131} \\
            \metricrule
            \quad + 1 Bezier      & \relv{ 3.000} & \relv{ 3.000}   & \tv[n]{77.81}{0.10} & \tv[n]{0.854}{0.001} & \tv[n]{0.012}{0.001} & \tv[n]{0.309}{0.000} & \tv[n]{\hs3.767}{0.137} \\
            \modelrule
            \gls{de}-3            & \relv{ 3.000} & \relv{ 3.000}   & \tv[n]{77.64}{0.04} & \tv[n]{0.862}{0.001} & \tv[n]{0.013}{0.001} & \tv[n]{0.311}{0.000} & \tv[n]{\hs3.000}{}      \\
            \metricrule
            \quad + 1 Bridge\brgs & \relv{ 3.086} & \relv{ 3.088}   & \tv[n]{77.77}{0.04} & \tv[n]{0.856}{0.001} & \tv[n]{0.012}{0.000} & \tv[n]{0.310}{0.000} & \tv[n]{\hs3.536}{0.043} \\
\bottomrule
        \end{tabular}
    }
    \label{tab:type2_in1k}
\end{table*}

\clearpage
\begin{table*}[t]
    \newcommand{\metricrule}{\cmidrule(lr){2-3} \cmidrule(lr){4-8}}
    \newcommand{\modelrule}{\cmidrule(lr){1-8}}
    \renewcommand{\arraystretch}{1.18}
    \centering
    \caption{
        Full result of performance improvement of the ensemble by adding type I bridges on CIFAR-10 dataset.
        We use the same settings as described in \cref{tab:type1} except that BatchNorm is applied instead of FRN.
    }
    \vspace{-2mm}
    \resizebox{0.9\textwidth}{!}{
        \begin{tabular}{lrrllllll}
            \multicolumn{8}{l}{\textbf{CIFAR-10 \ (w/ BatchNorm)}} \\
            \toprule
            Model                 & \dmet{FLOPs} & \dmet{\#Params} & \umet{ACC}          & \dmet{NLL}           & \dmet{ECE}           & \dmet{BS}            & \umet{DEE}           \\
            \midrule
            ResNet (\gls{de}-1)   & \relv{1.000} & \relv{1.000}    & \tv[n]{93.33}{0.17} & \tv[n]{0.212}{0.006} & \tv[n]{0.008}{0.002} & \tv[n]{0.100}{0.002} & \tv[n]{1.000}{}      \\
            \metricrule
            \quad + 1 Bridge\brgs & \relv{1.063} & \relv{1.048}    & \tv[n]{93.72}{0.03} & \tv[n]{0.192}{0.001} & \tv[b]{0.007}{0.001} & \tv[n]{0.093}{0.000} & \tv[n]{1.611}{0.085} \\
            \quad + 2 Bridge\brgs & \relv{1.125} & \relv{1.097}    & \tv[n]{93.77}{0.04} & \tv[n]{0.189}{0.001} & \tv[b]{0.007}{0.001} & \tv[b]{0.092}{0.000} & \tv[n]{1.696}{0.073} \\
            \quad + 3 Bridge\brgs & \relv{1.188} & \relv{1.145}    & \tv[b]{93.80}{0.07} & \tv[n]{0.188}{0.001} & \tv[n]{0.008}{0.001} & \tv[b]{0.092}{0.000} & \tv[n]{1.727}{0.095} \\
            \quad + 4 Bridge\brgs & \relv{1.250} & \relv{1.194}    & \tv[n]{93.77}{0.01} & \tv[b]{0.187}{0.001} & \tv[b]{0.007}{0.001} & \tv[b]{0.092}{0.000} & \tv[b]{1.749}{0.113} \\
            \metricrule
            \quad + 1 Bridge\brgm & \relv{1.207} & \relv{1.159}    & \tv[n]{93.70}{0.03} & \tv[n]{0.190}{0.001} & \tv[b]{0.007}{0.000} & \tv[n]{0.093}{0.000} & \tv[n]{1.654}{0.075} \\
\bottomrule
        \end{tabular}
    }
    \label{tab:type1_c10_bn}
\end{table*}

\begin{table*}[t]
    \newcommand{\metricrule}{\cmidrule(lr){2-3} \cmidrule(lr){4-8}}
    \newcommand{\modelrule}{\cmidrule(lr){1-8}}
    \renewcommand{\arraystretch}{1.18}
    \centering
    \caption{
        Full result of performance improvement of the ensemble by adding type II bridges on CIFAR-10 dataset.
        We use the same settings as described in \cref{tab:type2} except that BatchNorm is applied instead of FRN.
    }
    \vspace{-2mm}
    \resizebox{0.9\textwidth}{!}{
        \begin{tabular}{lrrlllll}
            \multicolumn{8}{l}{\textbf{CIFAR-10 \ (w/ BatchNorm)}} \\
            \toprule
            Model                 & \dmet{FLOPs}  & \dmet{\#Params} & \umet{ACC}          & \dmet{NLL}           & \dmet{ECE}           & \dmet{BS}            & \umet{DEE}           \\
            \midrule
            \gls{de}-2            & \relv{ 2.000} & \relv{ 2.000}   & \tv[n]{94.39}{0.04} & \tv[n]{0.179}{0.002} & \tv[n]{0.008}{0.002} & \tv[n]{0.086}{0.001} & \tv[n]{2.000}{}      \\
            \metricrule
            \quad + 1 Bridge\brgs & \relv{ 2.082} & \relv{ 2.063}   & \tv[b]{94.33}{0.05} & \tv[b]{0.174}{0.001} & \tv[b]{0.005}{0.001} & \tv[b]{0.085}{0.001} & \tv[b]{2.603}{0.056} \\
            \metricrule
            \quad + 1 Bridge\brgm & \relv{ 2.249} & \relv{ 2.192}   & \tv[b]{94.35}{0.13} & \tv[b]{0.174}{0.002} & \tv[b]{0.006}{0.001} & \tv[b]{0.084}{0.001} & \tv[b]{2.630}{0.090} \\
            \metricrule
            \quad + 1 Bezier      & \relv{ 3.000} & \relv{ 3.000}   & \tv[n]{94.28}{0.07} & \tv[n]{0.178}{0.003} & \tv[n]{0.006}{0.001} & \tv[n]{0.086}{0.001} & \tv[n]{2.136}{0.097} \\
            \modelrule
            \gls{de}-3            & \relv{ 3.000} & \relv{ 3.000}   & \tv[n]{94.56}{0.10} & \tv[n]{0.170}{0.001} & \tv[n]{0.007}{0.000} & \tv[n]{0.082}{0.001} & \tv[n]{3.000}{}      \\
            \metricrule
            \quad + 1 Bridge\brgs & \relv{ 3.082} & \relv{ 3.063}   & \tv[b]{94.51}{0.03} & \tv[n]{0.166}{0.001} & \tv[b]{0.005}{0.000} & \tv[b]{0.081}{0.001} & \tv[n]{3.672}{0.068} \\
\bottomrule
        \end{tabular}
    }
    \label{tab:type2_c10_bn}
\end{table*}

\clearpage
\begin{table*}[t]
    \newcommand{\metricrule}{\cmidrule(lr){2-3} \cmidrule(lr){4-8}}
    \newcommand{\modelrule}{\cmidrule(lr){1-8}}
    \renewcommand{\arraystretch}{1.18}
    \centering
    \caption{
        Full result of performance improvement of the ensemble by adding type I bridges on CIFAR-100 dataset.
        We use the same settings as described in \cref{tab:type1} except that BatchNorm is applied instead of FRN.
    }
    \vspace{-2mm}
    \resizebox{0.9\textwidth}{!}{
        \begin{tabular}{lrrlllll}
            \multicolumn{8}{l}{\textbf{CIFAR-100 \ (w/ BatchNorm)}} \\
            \toprule
            Model                 & \dmet{FLOPs} & \dmet{\#Params} & \umet{ACC}          & \dmet{NLL}           & \dmet{ECE}           & \dmet{BS}            & \umet{DEE}           \\
            \midrule
            ResNet (\gls{de}-1)   & \relv{1.000} & \relv{1.000}    & \tv[n]{76.48}{0.18} & \tv[n]{0.909}{0.009} & \tv[n]{0.036}{0.001} & \tv[n]{0.332}{0.003} & \tv[n]{1.000}{}      \\
            \metricrule
            \quad + 1 Bridge\brgs & \relv{1.063} & \relv{1.048}    & \tv[n]{77.45}{0.12} & \tv[n]{0.828}{0.007} & \tv[n]{0.021}{0.003} & \tv[n]{0.316}{0.002} & \tv[n]{1.718}{0.086} \\
            \quad + 2 Bridge\brgs & \relv{1.127} & \relv{1.097}    & \tv[n]{77.77}{0.10} & \tv[n]{0.805}{0.004} & \tv[n]{0.019}{0.001} & \tv[n]{0.310}{0.002} & \tv[n]{1.937}{0.088} \\
            \quad + 3 Bridge\brgs & \relv{1.190} & \relv{1.145}    & \tv[b]{77.87}{0.04} & \tv[n]{0.793}{0.002} & \tv[n]{0.017}{0.001} & \tv[n]{0.307}{0.001} & \tv[n]{2.062}{0.124} \\
            \quad + 4 Bridge\brgs & \relv{1.254} & \relv{1.194}    & \tv[n]{77.85}{0.11} & \tv[b]{0.788}{0.002} & \tv[b]{0.016}{0.001} & \tv[b]{0.306}{0.001} & \tv[b]{2.148}{0.128} \\
            \metricrule
            \quad + 1 Bridge\brgm & \relv{1.210} & \relv{1.160}    & \tv[n]{77.56}{0.16} & \tv[n]{0.819}{0.008} & \tv[n]{0.021}{0.001} & \tv[n]{0.313}{0.002} & \tv[n]{1.798}{0.102} \\
\bottomrule
        \end{tabular}
    }
    \label{tab:type1_c100_bn}
\end{table*}

\begin{table*}[t]
    \newcommand{\metricrule}{\cmidrule(lr){2-3} \cmidrule(lr){4-8}}
    \newcommand{\modelrule}{\cmidrule(lr){1-8}}
    \renewcommand{\arraystretch}{1.18}
    \centering
    \caption{
        Full result of performance improvement of the ensemble by adding type II bridges on CIFAR-100 dataset.
        We use the same settings as described in \cref{tab:type2} except that BatchNorm is applied instead of FRN.
    }
    \vspace{-2mm}
    \resizebox{0.9\textwidth}{!}{
        \begin{tabular}{lrrlllll}
            \multicolumn{8}{l}{\textbf{CIFAR-100 \ (w/ BatchNorm)}} \\
            \toprule
            Model                 & \dmet{FLOPs}  & \dmet{\#Params} & \umet{ACC}          & \dmet{NLL}           & \dmet{ECE}           & \dmet{BS}            & \umet{DEE}           \\
            \midrule
            \gls{de}-2            & \relv{ 2.000} & \relv{ 2.000}   & \tv[n]{78.73}{0.09} & \tv[n]{0.795}{0.005} & \tv[n]{0.021}{0.003} & \tv[n]{0.299}{0.001} & \tv[n]{2.000}{}      \\
            \metricrule
            \quad + 1 Bridge\brgs & \relv{ 2.083} & \relv{ 2.063}   & \tv[b]{79.15}{0.13} & \tv[b]{0.757}{0.002} & \tv[b]{0.015}{0.001} & \tv[b]{0.292}{0.001} & \tv[b]{2.812}{0.075} \\
            \metricrule
            \quad + 1 Bridge\brgm & \relv{ 2.253} & \relv{ 2.193}   & \tv[b]{79.22}{0.12} & \tv[b]{0.755}{0.002} & \tv[b]{0.015}{0.001} & \tv[b]{0.291}{0.001} & \tv[b]{2.872}{0.088} \\
            \metricrule
            \quad + 1 Bezier      & \relv{ 3.000} & \relv{ 3.000}   & \tv[n]{79.17}{0.13} & \tv[n]{0.765}{0.002} & \tv[n]{0.017}{0.001} & \tv[n]{0.292}{0.001} & \tv[n]{2.661}{0.111} \\
            \modelrule
            \gls{de}-3            & \relv{ 3.000} & \relv{ 3.000}   & \tv[n]{79.79}{0.13} & \tv[n]{0.748}{0.005} & \tv[n]{0.020}{0.001} & \tv[n]{0.285}{0.001} & \tv[n]{3.000}{}      \\
            \metricrule
            \quad + 1 Bridge\brgs & \relv{ 3.083} & \relv{ 3.063}   & \tv[n]{79.93}{0.11} & \tv[n]{0.728}{0.003} & \tv[n]{0.016}{0.001} & \tv[n]{0.282}{0.001} & \tv[n]{3.847}{0.172} \\
\bottomrule
        \end{tabular}
    }
    \label{tab:type2_c100_bn}
\end{table*}

\clearpage
\begin{table*}[t]
    \newcommand{\metricrule}{\cmidrule(lr){2-3} \cmidrule(lr){4-8}}
    \newcommand{\modelrule}{\cmidrule(lr){1-8}}
    \renewcommand{\arraystretch}{1.18}
    \centering
    \caption{
        Full result of performance improvement of the ensemble by adding type I bridges on Tiny ImageNet dataset.
        We use the same settings as described in \cref{tab:type1} except that BatchNorm is applied instead of FRN.
    }
    \vspace{-2mm}
    \resizebox{0.9\textwidth}{!}{
        \begin{tabular}{lrrlllll}
            \multicolumn{8}{l}{\textbf{Tiny ImageNet \ (w/ BatchNorm)}} \\
            \toprule
            Model                 & \dmet{FLOPs} & \dmet{\#Params} & \umet{ACC}          & \dmet{NLL}           & \dmet{ECE}           & \dmet{BS}            & \umet{DEE}           \\
            \midrule
            ResNet (\gls{de}-1)   & \relv{1.000} & \relv{1.000}    & \tv[n]{63.90}{0.11} & \tv[n]{1.560}{0.006} & \tv[n]{0.035}{0.004} & \tv[n]{0.483}{0.002} & \tv[n]{1.000}{}      \\
            \metricrule
            \quad + 1 Bridge\brgs & \relv{1.050} & \relv{1.057}    & \tv[n]{66.24}{0.12} & \tv[n]{1.378}{0.006} & \tv[n]{0.018}{0.004} & \tv[n]{0.446}{0.001} & \tv[n]{2.354}{0.113} \\
            \quad + 2 Bridge\brgs & \relv{1.099} & \relv{1.114}    & \tv[n]{67.34}{0.06} & \tv[n]{1.332}{0.005} & \tv[n]{0.017}{0.002} & \tv[n]{0.436}{0.000} & \tv[n]{3.130}{0.132} \\
            \quad + 3 Bridge\brgs & \relv{1.149} & \relv{1.171}    & \tv[n]{67.77}{0.13} & \tv[n]{1.313}{0.003} & \tv[n]{0.016}{0.001} & \tv[n]{0.431}{0.000} & \tv[n]{3.678}{0.106} \\
            \quad + 4 Bridge\brgs & \relv{1.198} & \relv{1.228}    & \tv[b]{67.94}{0.11} & \tv[b]{1.300}{0.002} & \tv[b]{0.016}{0.001} & \tv[b]{0.429}{0.000} & \tv[b]{4.050}{0.120} \\
            \metricrule
            \quad + 1 Bridge\brgm & \relv{1.180} & \relv{1.206}    & \tv[n]{66.61}{0.12} & \tv[n]{1.358}{0.006} & \tv[n]{0.016}{0.002} & \tv[n]{0.441}{0.001} & \tv[n]{2.675}{0.132} \\
\bottomrule
        \end{tabular}
    }
    \label{tab:type1_t200_bn}
\end{table*}

\begin{table*}[t]
    \newcommand{\metricrule}{\cmidrule(lr){2-3} \cmidrule(lr){4-8}}
    \newcommand{\modelrule}{\cmidrule(lr){1-8}}
    \renewcommand{\arraystretch}{1.18}
    \newcommand{\hs}{\hspace{1.8mm}}
    \centering
    \caption{
        Full result of performance improvement of the ensemble by adding type II bridges on Tiny ImageNet dataset.
        We use the same settings as described in \cref{tab:type2} except that BatchNorm is applied instead of FRN.
    }
    \vspace{-2mm}
    \resizebox{0.9\textwidth}{!}{
        \begin{tabular}{lrrlllll}
            \multicolumn{8}{l}{\textbf{Tiny ImageNet \ (w/ BatchNorm)}} \\
            \toprule
            Model                 & \dmet{FLOPs}  & \dmet{\#Params} & \umet{ACC}          & \dmet{NLL}           & \dmet{ECE}           & \dmet{BS}            & \umet{DEE}           \\
            \midrule
            \gls{de}-2            & \relv{ 2.000} & \relv{ 2.000}   & \tv[n]{66.92}{0.22} & \tv[n]{1.401}{0.005} & \tv[n]{0.024}{0.003} & \tv[n]{0.444}{0.002} & \tv[n]{\hs2.000}{}      \\
            \metricrule
            \quad + 1 Bridge\brgs & \relv{ 2.059} & \relv{ 2.067}   & \tv[b]{68.13}{0.27} & \tv[b]{1.315}{0.007} & \tv[b]{0.020}{0.002} & \tv[b]{0.427}{0.002} & \tv[b]{\hs3.616}{0.227} \\
            \metricrule
            \quad + 1 Bridge\brgm & \relv{ 2.198} & \relv{ 2.227}   & \tv[b]{68.19}{0.12} & \tv[b]{1.305}{0.005} & \tv[b]{0.018}{0.002} & \tv[b]{0.425}{0.001} & \tv[b]{\hs3.898}{0.208} \\
            \metricrule
            \quad + 1 Bezier      & \relv{ 3.000} & \relv{ 3.000}   & \tv[n]{68.74}{0.17} & \tv[n]{1.306}{0.006} & \tv[n]{0.021}{0.002} & \tv[n]{0.420}{0.002} & \tv[n]{\hs3.873}{0.255} \\
            \modelrule
            \gls{de}-3            & \relv{ 3.000} & \relv{ 3.000}   & \tv[n]{68.21}{0.17} & \tv[n]{1.337}{0.006} & \tv[n]{0.023}{0.001} & \tv[n]{0.427}{0.001} & \tv[n]{\hs3.000}{}      \\
            \metricrule
            \quad + 1 Bridge\brgs & \relv{ 3.059} & \relv{ 3.067}   & \tv[n]{68.94}{0.24} & \tv[n]{1.282}{0.005} & \tv[n]{0.022}{0.001} & \tv[n]{0.418}{0.001} & \tv[n]{\hs4.773}{0.173} \\
\bottomrule
        \end{tabular}
    }
    \label{tab:type2_t200_bn}
\end{table*}

\end{document}